\documentclass[letterpaper]{article} 
\usepackage{aaai2026}  
\usepackage{times}  
\usepackage{helvet}  
\usepackage{courier}  
\usepackage[hyphens]{url}  
\usepackage{graphicx} 
\usepackage{natbib}  
\usepackage{caption} 
\usepackage{algorithm}
\usepackage{algorithmic}

\usepackage{newfloat}
\usepackage{listings}
\DeclareCaptionStyle{ruled}{labelfont=normalfont,labelsep=colon,strut=off} 
\floatstyle{ruled}
\newfloat{listing}{tb}{lst}{}
\floatname{listing}{Listing}
\def\ie{{\textit{i.e. }}}

\def\etal{{\textit{et al. }}}

\newcommand{\blux}[1]{{\color{blue}#1}}

\usepackage{amsmath}
\usepackage{amssymb}
\usepackage{booktabs}
\usepackage{multirow}
\usepackage{autobreak}
\usepackage{makecell}
\usepackage{xcolor} 
\usepackage{colortbl} 
\definecolor{Gray}{gray}{0.9}

\usepackage{multicol}
\usepackage[most]{tcolorbox}

\lstdefinestyle{promptstyle}{
    basicstyle=\ttfamily\tiny,  
    breaklines=true,
    numbers=none,
    keepspaces=true,
    columns=flexible,
    backgroundcolor=\color{gray!5},
    frame=none
}

\newtcolorbox{promptbox}[1][]{
    colback=white,
    colframe=gray!50,
    boxrule=0.5pt,
    arc=3pt,
    outer arc=3pt,
    leftrule=2pt,
    rightrule=2pt,
    toprule=2pt,
    bottomrule=2pt,
    left=5pt,
    right=5pt,
    top=5pt,
    bottom=5pt,
    enhanced,
    shadow={2pt}{-2pt}{0pt}{black!20}
}

\usepackage{cleveref}
\crefname{figure}{Fig.}{Figs.}
\crefname{table}{Tab.}{Tables}
\crefname{equation}{Eq.}{Eqs.}
\crefname{section}{Sec.}{Secs.}
\crefname{algorithm}{Algorithm.}{Algorithm}

\title{Rectification Reimagined: A Unified Mamba Model for Image Correction and Rectangling with Prompts}
\author{
    Linwei Qiu\textsuperscript{\rm 1,\rm 2,\rm 3,\rm 4},
    Gongzhe Li\textsuperscript{\rm 5},
    Xiaozhe Zhang\textsuperscript{\rm 1,\rm 2,\rm 3},
    Qilin Sun\textsuperscript{\rm 5},
    Fengying Xie\textsuperscript{\rm 1,\rm 2,\rm 3}\thanks{Corresponding Author}
}
\affiliations{
    \textsuperscript{\rm 1}Tianmushan Laboratory, Beihang University, Hangzhou, China\\
    \textsuperscript{\rm 2}State Key Laboratory of High-Efficiency Reusable Aerospace Transportation Technology, Beijing, China\\
    \textsuperscript{\rm 3}School of Astronautics, Beihang University, Beijing, China\\
    \textsuperscript{\rm 4}School of Aerospace Engineering, Huazhong University of Science and Technology, Wuhan, China\\
    \textsuperscript{\rm 5}School of Data Science, The Chinese University of Hong Kong, Shenzhen\\
        \{qiulinwei, xiaozhe\_zhang, xfy\_73\}@buaa.edu.cn, gongzheli1@link.cuhk.edu.cn, sunqilin@cuhk.edu.cn

}

\usepackage{bibentry}
\begin{document}

\maketitle

\begin{figure*}[t!]
    \centering
     \includegraphics[width=\linewidth]{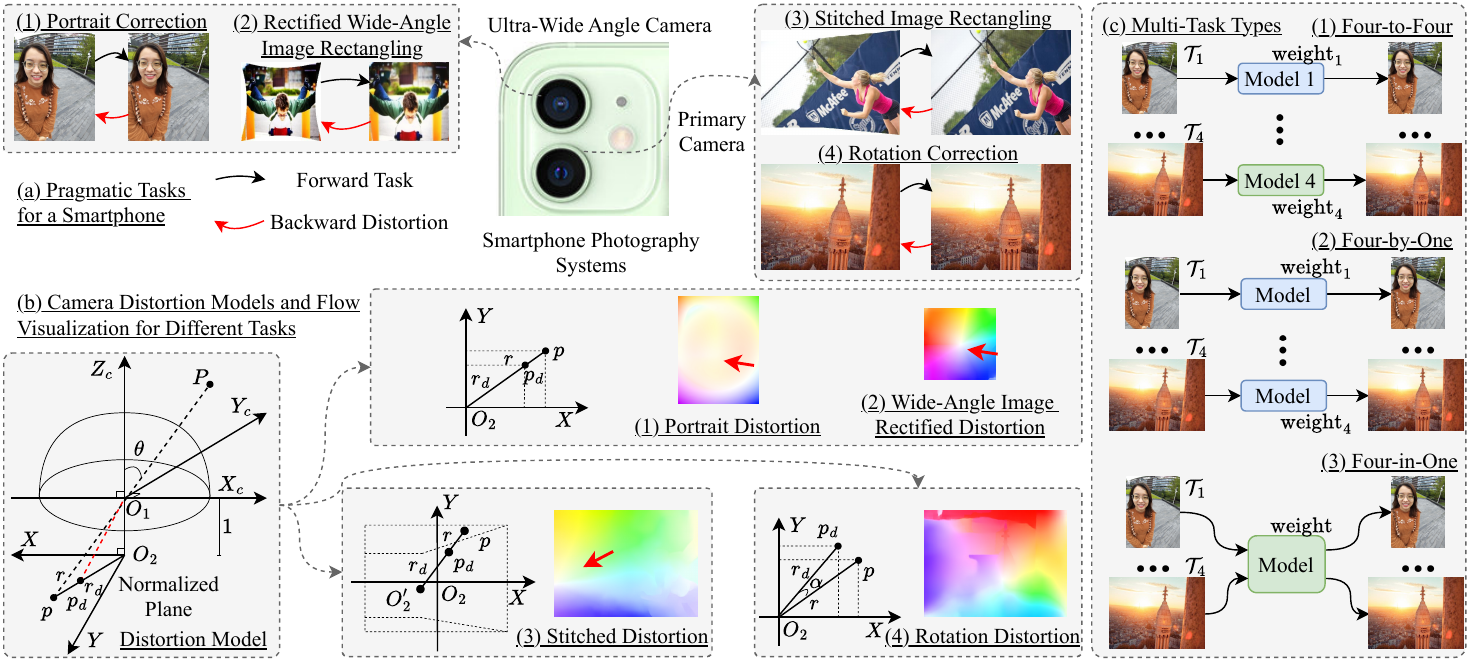}
    \caption{\underline{(a) Pragmatic tasks for a smartphone.} We consider four detailed tasks for image correction and rectangling, which are closely related to two types of common cameras on a mainstream mobile phone. Portrait correction ($\mathcal{T}_1$) and rectified wide-angle image rectangling ($\mathcal{T}_2$) often need to process pictures taken with a wide-angle lens. Stitched image rectangling ($\mathcal{T}_3$) and rotation correction ($\mathcal{T}_4$) are two practical tasks for daily life. 
    \underline{(b) Camera distortion models and flow visualization for different tasks.} On a normalized plane, the image of the point $P$ is $p_d$ whereas it would be $p$ without distortion following a pinhole camera model \cite{kannala2006generic}. The optical flows of backward distortions are generated through the RAFT \cite{teed2020raft}, which is the one of the most powerful tools for optical flow estimation. 
    \underline{(c) Multi-task types.} (1) four-to-four. Previous studies often use four disparate models to accomplish four tasks separately. (2) four-by-one. four tasks are achieved by one model structure but sharing different even counteractive network weights. (3) four-in-one. four tasks are employed in one model which can handle diverse tasks at the same time. four-by-one and four-in-one are all task-agnostic models. 
    We obtain the \textbf{four-by-one} and \textbf{four-in-one} in this paper.
    }
    \label{fig:intro}
\end{figure*}

\begin{abstract}
Image correction and rectangling are valuable tasks in practical photography systems such as smartphones.
Recent remarkable advancements in deep learning have undeniably brought about substantial performance improvements in these fields. Nevertheless, existing methods mainly rely on task-specific architectures. This significantly restricts their generalization ability and effective application across a wide range of different tasks.
In this paper, we introduce the \textbf{Unified Rectification Framework (UniRect)}, a comprehensive approach that addresses these practical tasks from a consistent distortion rectification perspective.
Our approach incorporates various task-specific inverse problems into a general distortion model by simulating different types of lenses.
To handle diverse distortions, UniRect adopts one task-agnostic rectification framework with a dual-component structure: a {Deformation Module}, which utilizes a novel Residual Progressive Thin-Plate Spline (RP-TPS) model to address complex geometric deformations, and a subsequent {Restoration Module}, which employs Residual Mamba Blocks (RMBs) to counteract the degradation caused by the deformation process and enhance the fidelity of the output image.
Moreover, a Sparse Mixture-of-Experts (SMoEs) structure is designed to circumvent heavy task competition in multi-task learning due to varying distortions. 
Extensive experiments demonstrate that our models have achieved state-of-the-art performance compared with other up-to-date methods. 
\end{abstract}

\begin{links}
    \link{Code}{https://github.com/yyywxk/UniRect}
\end{links}

\section{Introduction}
\label{sec:intro}
With the rapid development of photography system, there has been a surging interest in low-level tasks, which play pivotal roles in enhancing image quality. 
Image correction usually aims at estimating and correcting distortions \cite{li2019blind, zhao2018distortion}, and has a very wide range of applications \cite{zhou2018gridface, zhuang2019learning, xue2019learning, zhan2019esir}. 
Image rectangling is a special type of post-processing task \cite{he2013rectangling} to acquire user-friendly rectangular images for photos with irregular boundaries while maintaining high content fidelity. 
Previous studies have addressed these problems by devising diverse specific architectures to learn different types of processing.
For instance, \citet{tan2021practical} proposed a cascaded network and progressively correct portrait distortion. \citet{nie2022deep} optimized a mesh model for stitched image rectangling. 
Nearly all of these methods concentrate on a single task and depend on task-specific networks. Besides, we note that these practical tasks may be required within one smartphone device and associated with specific cameras or lenses, as depicted in \cref{fig:intro}a. 
The former indicates that identifying a unified task-agnostic model could be highly beneficial for edge devices with limited computation capabilities and memory resources.
The latter suggests that potential distortion relationships may exist in other tasks beyond the portrait and rotation correction tasks when considering all tasks as one general rectification task.

This work considers these problems from \textbf{a consistent rectification perspective}.
We rethink the inverse processes of these problems as the distortion and demonstrate that they can be encompassed within a general distortion model in \cref{fig:intro}b. 
Here, rectified wide-angle image rectangling and stitched image rectangling can be viewed as a unified rectification task addressing {wide-angle image rectified distortion and {stitched distortion} respectively. These two types of distortion lead to irregular boundaries, which may impair image quality and vision understanding. 
Subsequently, we propose a unified rectification framework (UniRect) to estimate such unified distortion and accomplish image correction and rectangling missions (including four specific tasks in \cref{fig:intro}a and obtaining four-by-one in \cref{fig:intro}c.2) by rectifying the input image. It consists of two main components: a deformation module based on residual progressive thin-plate spline (RP-TPS) model, which approximates the distortion by progressively predicting the locations of control points; a restoration module based on Residual Mamba Blocks (RMBs) to counteract the degradation caused by previous deformation process. 
Moreover, to address the heavy task competition in multi-task learning, we design a Sparse Mixture-of-Experts (SMoEs) strategy and perform four-in-one in \cref{fig:intro}c.3.
This approach uses a gating network to assign weights to expert networks, thereby alleviating performance degradation.  Our models have attained state-of-the-art performance levels comparable to those of current methods across four task scenarios.

In summary, our main contributions are as follows:
\begin{itemize}
    \item We take a novel view of image correction and rectangling from a consistent rectification perspective and establish a general distortion model mathematically to describe their backward distortion processes.
    \item We propose a {Unified Rectification Framework (UniRect)} with prompts to effectively address four typical tasks ({four-by-one}) and design a Sparse Mixture-of-Experts (SMoEs) structure for multi-task learning ({four-in-one}).
    \item We conduct extensive experiments on various datasets to validate the superior of our models compared to up-to-date methods across different tasks.
\end{itemize}

\section{Related Work}
\label{sec:related}
\textbf{Image Correction.}
Portrait distortion is a common issue encountered in smartphone photography. \citet{shih2019distortion} proposed an algorithm based on mesh optimization to restore the wrapped content. \citet{tan2021practical} introduced the first end-to-end method based on deep learning and provided a supervised dataset and \citet{zhu2022semi} proposed a semi-supervised method to further improve results.
Rotation distortion is caused by arbitrary rotation angle. Direct rotation requires angle-prior and also destroys the rectangular boundaries. Some previous methods preserve image structure (edge or line) to ensure the naturalness of image content \cite{he2013content, von2008lsd}.
\citet{nie2023deep} proposed a mesh transformation network to correct content and boundaries. However, all these networks need task-specific designs.

\noindent\textbf{Image Rectangling.}
To obtain rectangular images after stitching, simple inner rectangle clipping leads to the loss of pixel semantics, while certain image inpainting or completion methods \cite{suvorov2022resolution, teterwak2019boundless, liao2021towards} pose the risk of introducing uncertain information. \citet{he2013rectangling} first introduced a content-aware algorithm that utilizes an energy function to optimize local mesh deformation.
Recent DNN-based methods \cite{nie2022deep, zhou2024recdiffusion, qiu2024remote} estimated the wrapping meshes or motion fields to handle non-rectangular borders. 
\citet{liao2023recrecnet} constructed a win-win representation for rectified wide-angle image rectangling. 
In this work, we regard these tasks as a unified rectification task.

\noindent\textbf{Multi-Task Learning.}
Multi-task learning methods can be roughly divided into task balancing and multi-task architecture.
Tasking balancing \cite{kendall2018multi, guo2018dynamic} methods focus on weights on task and gradient conflicts by re-weighting the loss or manipulating the gradient.
This type of method can successfully address multi-task learning for similar problems \cite{liao2024mowa}, such as simultaneous image denoising and deblurring \cite{zamir2021multi, potlapalli2024promptir}. However, it tends to underperform when there is a significant disparity between different tasks.
Besides, some other approaches \cite{bragman2019stochastic,ruder2019latent, gao2019nddr} propose the advanced architecture for parameter sharing.
\citet{gao2019nddr} employed different decoders for different tasks, while \citet{ruder2019latent} proposed a gating mechanism as soft parameters sharing between different task networks. We try to mitigate task competition and achieve the four-in-one in this paper.

\begin{figure*}[t!]
    \centering
     \includegraphics[width=\linewidth]{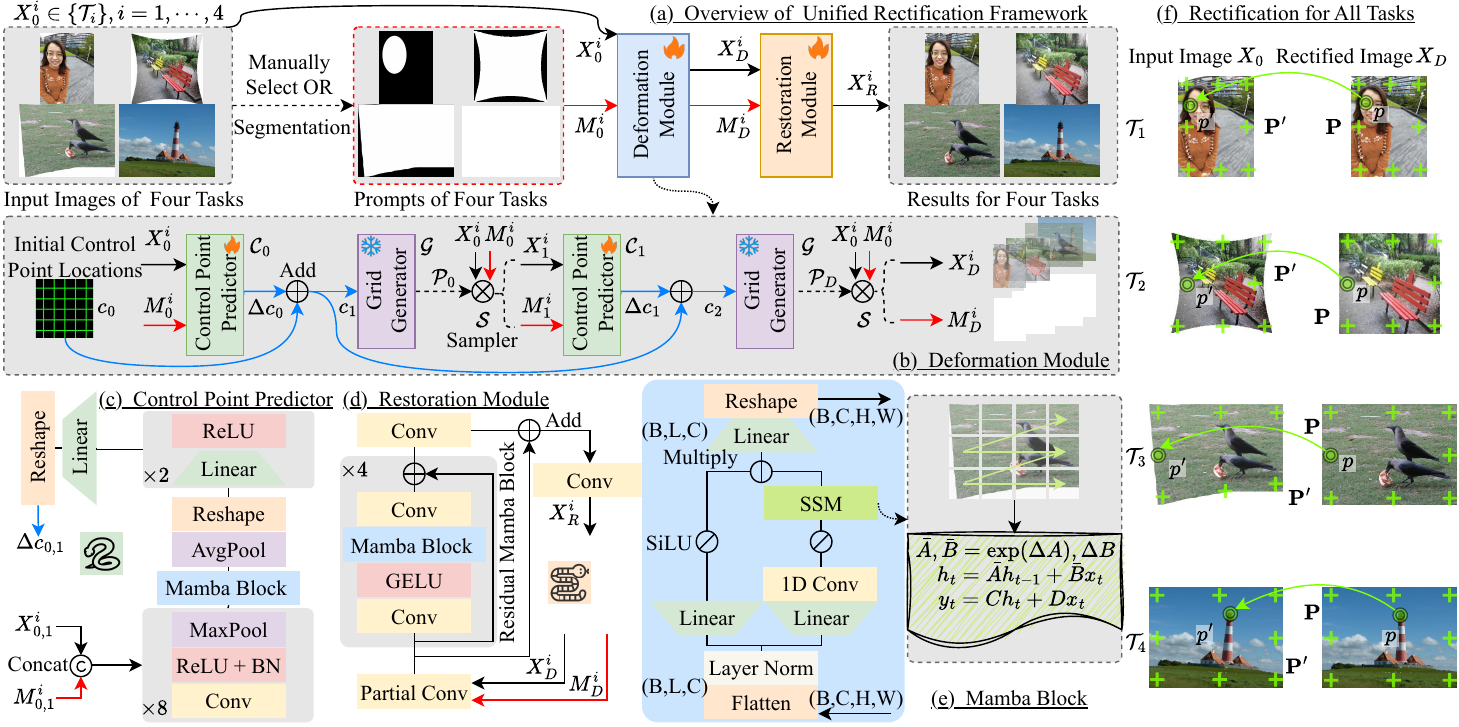}
    \caption{
    (a) Framework of our Unified Rectification. It mainly consists of a (b) deformation module (DM) and a (d) restoration module (RM), which are trained simultaneously. An image $X^i_0$ from the image set of task $\mathcal{T}_i$ and a corresponding visual prompt $M^i_0$ indicating which task to perform are set as input for UniRect, which yields the final result $X^i_R$. In deformation module (ignoring our residual progressive setting for simplicity), the (c) control point predictor $\mathcal{C}$ can predict the locations of a set of control points, \ie $c$ , with which the grid generator $\mathcal{G}$ produces a sampling grid $\mathcal{P}$ by \cref{eq:5}. The sampler $\mathcal{S}$ then samples from $X^i_0$ and $M^i_0$ with the restriction of $\mathcal{P}$, resulting in the rectified image $X^i_D$ and its new prompt $M^i_D$. For some tasks without boundary changes like $\mathcal{T}_1$ and $\mathcal{T}_4$, $M^i_D$ will be changed into the all-one matrix. (e) Mamba block is applied in the control point predictor to obtain geometric information like borders by its scan characteristic \cite{gu2021combining, gu2023mamba}. RM is composed of residual mamba blocks (RMBs), which effectively captures global connections for restoration. 
    (f) Rectification for all tasks. Our model treats all tasks as rectification tasks and subsequently incorporates them into a unified framework.
    }
    \label{fig:arc}
\end{figure*}

\begin{figure*}[t]
    \centering
    \includegraphics[width=\linewidth]{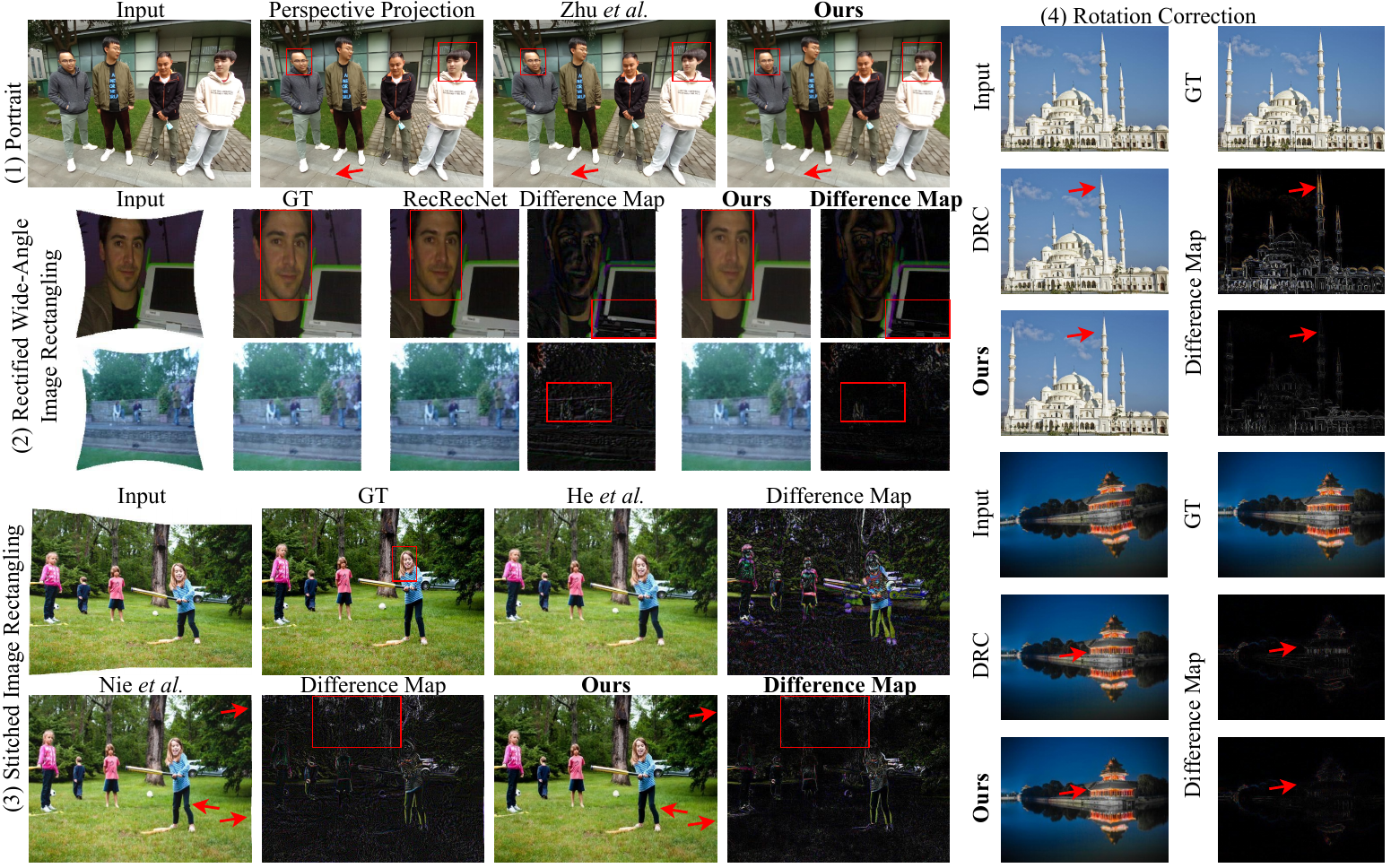}
    \caption{
    Qualitative comparison for our UniRect on four tasks. Zoom in for best view.
    }
    \label{fig:compare}
\end{figure*}

\section{Methodology}
\label{sec:method}

\subsection{General Distortion Model}
We initiate the construction of distortion models for these tasks by leveraging optical flow. The inverse problems $\mathcal{T}^{-1}$ of these tasks $\mathcal{T}$ are backward distortion procedures, which are depicted in \cref{fig:intro}b. $X_cO_1Y_c$ is a camera coordinate system and $XO_2Y$ is the scaled coordinate system. For the image point $p=[x,y]^T$ and its distortion point $p_d=[x_d,y_d]^T$, $r$ and $r_d$ are distances between them and the principal point $O_2$ respectively. We categorize the backward distortion into four types, based on tasks and different flow visualization maps. 

\underline{\textbf{1)}} For portrait distortion in $\mathcal{T}_1^{-1}$, it has been fully studied by previous researchers and the Kanala-Brandt model \cite{kannala2006generic} is widely used to approximate the distortion of a wide-angle lens: $r_d = \sum_{j = 1}^N {k_j{\theta^{2j - 1}}}, N=1,2,3,\cdots$, where $\theta$ denotes the angle between the incoming ray and the optical axis and $k_j$ is the distortion parameter. 
\underline{\textbf{2)}} For wide-angle image rectified distortion in $\mathcal{T}_2^{-1}$, we observe that its flow appears a radial distribution as indicated by the red arrow in \cref{fig:intro}b.2 and a traditional Brown-Conrady model \cite{weng1992camera} in introduced here: $r_d = \sum_{j = 1}^N {k_j{r^{2j - 1}}}, N=1,2,3,\cdots$. 
\underline{\textbf{3)}} The stitched distortion in $\mathcal{T}^{-1}_3$ is formed by supposing that the rectangular image is transmuted into a image with irregular boundaries through an amorphous and vibrant lens. We discover that the flow in \cref{fig:intro}b.3 presents an approximate radial distribution around a particular point $O_2^{\prime}$, which is affected by irregular borders. Thus, we biuld a variant of Brown-Conrady model to describe this phenomenon:

\begin{footnotesize}
\begin{equation}\label{eq:1}
    \left[ {\begin{array}{*{20}{c}}
{{x_d}}\\
{{y_d}}
\end{array}} \right] = \frac{{{r_d}}}{r} \left[ {\begin{array}{*{20}{c}}
x\\
y
\end{array}} \right] + {T}_0 = \frac{1}{r} \left( {\sum\limits_{j = 1}^N {k_j{{r}^{2j - 1}}} } \right) \left[ {\begin{array}{*{20}{c}}
x\\
y
\end{array}} \right] + {T}_0, 
\end{equation}
\end{footnotesize}
where ${T}_0=[x_0,y_0]^T$ penalizes the decentered influence of $O_2^{\prime}$. 
\underline{\textbf{4)}} The rotation distortion in $\mathcal{T}^{-1}_4$ need to consider the rotation angle $\alpha$ of the coordinates:

\begin{footnotesize}
\begin{equation}\label{eq:2}
\left[ {\begin{array}{*{20}{c}}
{{x_d}}\\
{{y_d}}
\end{array}} \right] = \frac{{{r_d}}}{r} \mathcal{R}_{\alpha} \left[ {\begin{array}{*{20}{c}}
x\\
y
\end{array}} \right]  
= \frac{1}{r} \left( {\sum\limits_{i = 1}^N {k_j{{r}^{2j - 1}}} } \right) \mathcal{R}_{\alpha} \left[ {\begin{array}{*{20}{c}}
x\\
y
\end{array}} \right], 
\end{equation}
\end{footnotesize}
where $\mathcal{R}_{\alpha}$ is the rotation matrix.

Considering all of the aforementioned, the final general distortion model is shown below

\begin{footnotesize}
\begin{equation}\label{eq:3}
\left[ {\begin{array}{*{20}{c}}
{{x_d}}\\
{{y_d}}
\end{array}} \right] 
 = \frac{1}{r} {\sum\limits_{j = 1}^N \left( {k_j{\theta ^{2j - 1}}}  + {k_j^{\prime}{r^{2j - 1}}}  \right) }\mathcal{R}_{\alpha} \left[ {\begin{array}{*{20}{c}}
x\\
y
\end{array}} \right] 
+ {T}_0, 
\end{equation}
\end{footnotesize}
where $k_j, k^{\prime}_j$ are distortion parameters. 
From the unified distortion model in \cref{eq:3}, we discover that the backward distortion of some tasks has a specific coupling relationship. 

\subsection{Framework of UniRect}
By describing inverse problems from a unified distortion perspective, we can view all these tasks as a unified distortion rectification task.
Our UniRect is depicted in \cref{fig:arc}. Here, the Deformation Module (DM) is designed to estimate task-specific distortions and carry out rectification. Concurrently, the Restoration Module (RM) is employed to compensate for potential degradation resulting from estimation errors and interpolation within the DM.

\noindent\textbf{Residual Progressive Thin-Plate Spline Model.}
As shown in \cref{eq:3}, this unified distortion process is inconstant and usually non-linear, which cannot be approximated by simple transformations. Therefore, we utilize Thin-Plate Spline (TPS) model to simulate aforementioned distortion. Give a set of basic control points $\mathbf{P}=\{p_1, \cdots, p_{N_c}\}$ on a rectified image and their corresponding control points $\mathbf{P}^{\prime}=\{p_1^{\prime}, \cdots, p_{N_c}^{\prime}\}$ on an image with distortion in \cref{fig:arc}f, the minimum approximation of TPS is formulated \cite{bookstein1989}:

\begin{footnotesize}
\begin{equation}\label{eq:4}
\begin{split}
    \min \;\sum\limits_{j = 1}^{{N_c}} {\left\| {\Phi \left( {{p_j}} \right) - {p^\prime }_j} \right\|}  
    &+ \lambda \iint_{\mathbb{R}^2} \left( {{\left( {\frac{{\partial \Phi }}{{\partial {x^2}}}} \right)}^2} \right.\\
    &\left. + 2{{\left( {\frac{{{\partial ^2}\Phi }}{{\partial x\partial y}}} \right)}^2}
    + {{\left( {\frac{{\partial \Phi }}{{\partial {y^2}}}} \right)}^2} \right)^2dxdy,
\end{split}
\end{equation}
\end{footnotesize}
where $\Phi$ denotes the transformation and $\lambda$ is a weight to balance two terms. The only closed-form solution \cite{kent1994link} of \cref{eq:4} can be derived as 

\begin{footnotesize}
    \begin{equation}\label{eq:5}
        \Phi \left( p \right) = A_{\Phi}{\left[ {\begin{array}{*{20}{c}}
p&1
\end{array}} \right]^T} + \sum\limits_{j = 1}^{{N_c}} {{w_j}U\left(\left\| {p - {p_j}} \right\|\right)},
    \end{equation}
\end{footnotesize}
where $p$ is any point on the rectified image and $U$ is the radial basis function. $A_\Phi \in \mathbb{R}^{2\times 3}, w_i \in \mathbb{R}^{2\times 1}$ are parameters which can be calculated by control point pairs. Therefore, if the basic control points are set to be evenly distributed throughout the rectified image, we can compute them from the distortion image by \cref{eq:5} so long as the locations of corresponding control points are given. This process and related matrices are fixed when the image resolution and $N_c$ are preset.

However, we discover that simply applying the TPS model does not perform well on some tasks. By introducing the initial control point locations $c_0$, we propose the residual progressive TPS (RP-TPS) model to progressively enhance the approximation and avoid intermediate interpolation errors. The flow of our deformation module based on RP-TPS is exhibited in \cref{fig:arc}b. Our initial control point locations $c_0$ are equal to our pre-defined basic control points. Two control point predictors $\mathcal{C}_0,\mathcal{C}_1$ with same structures are designed to predict the deviations of control points based on their inputs. In summary, the overall procedures through RP-TPS are formulated by

\begin{footnotesize}
    \begin{equation}\label{eq:6}
     {c_1} = {c_0} + {\mathcal{C}_0}\left( {X_0^i,M_0^i} \right),X_1^i = \mathcal{S}\left[ {\mathcal{G}\left( {{c_1}} \right);X_0^i} \right],
    \end{equation}
    \begin{equation}\label{eq:7}
     {c_2} = {c_1} + {\mathcal{C}_1}\left( {X_1^i,M_1^i} \right),X_D^i = \mathcal{S}\left[ {\mathcal{G}\left( {{c_2}} \right);X_0^i} \right],
    \end{equation}
\end{footnotesize}
where $\mathcal{G}$ is a grid generator which produces a grid $\mathcal{P}$ on the coordinate space of $X_0^i$ with distortions by iterating all points in the rectified image. $\mathcal{S}$ is the differentiable bilinear sampler. Sampling processes in \cref{eq:6} and \cref{eq:7} are only from the foremost inputs $X_0^i$ and visual prompts $M_0^i$, which effectively alleviates the interpolation degradation of the intermediate process. 

\noindent\textbf{Prompt Designs.} We elaborate different visual prompts for different tasks, which is also a reflection of \cref{eq:3} and helps the network focus on task-specific distortion. Specifically, for $\mathcal{T}_1$, $M_0^1$ is the face mask since the face distortion is import for a portrait photo in this task. $M_0^2$ and $M_0^3$ indicate the irregular boundaries, thereby enhancing the perception of these areas, and they also contribute to the loss function (\cref{eq:8}). For rotation correction, the prompt $M_0^4$ is a white image as the model should pay attention to the whole content of the input image to perceive the subtle rotation distortion.

\noindent\textbf{Network Structures.} 
\label{sec:net}
The structure of Control Point Predictor $\mathcal{C}_{0,1}$ is shown in \cref{fig:arc}c. The image $X^i_{0,1}$ and its prompt $M^i_{0,1}$ are concatenated and fed into $\mathcal{C}_{0,1}$ to generate the deviation locations $\Delta c_{0,1} \in \mathbb{R}^{B\times N_c \times 2}$ of control points $P^\prime$ in $X^i_{0,1}$. Inspired by recent burgeoning development of Mamba \cite{gu2021combining,gu2023mamba}, a mamba block is introduced in the latent space to scan the geometric information and handle insidious long-range dependencies of distortions. 
The structure of {Restoration Module} is shown in \cref{fig:arc}d, which is mainly composed of four Residual Mamba Blocks (RMBs). In practice, each RMB has $32$ channels in convolution layers and a same mamba block is in the core of RMB to capture non-local connections effectively like many vision tasks \cite{liu2024vmamba, zhu2024vision, guo2025mambair}. 
$X^i_D$ and $M^i_D$ first pass through a partial convolution layer \cite{Liu2018ImageIF} as there may be some irregular borders in $X^i_D$ for boundary-changing task like $\mathcal{T}_2, \mathcal{T}_3, \mathcal{T}_3$ due to the inaccurate estimation of the positions of the control points in the previous module. For $\mathcal{T}_1, \mathcal{T}_4$, the partial convolution is equivalent to a standard convolution layer since the $M^i_D$ is set to all-one matrix. 

\subsection{Loss Functions}
\label{sec:loss}
For deformation module, we propose a boundary loss $\mathcal{L}_b$ to facilitate the outermost points $P^{\prime}+$ of the control points $P^{\prime}$ approaching the boundary. 
Given a prompt $M$ indicating boundaries (for $\mathcal{T}_2,\mathcal{T}_3,\mathcal{T}_3$), the $\mathcal{L}_b$ can be calculated by $\mathcal{L}_b = \sum_{p^{\prime} \in P^{\prime}+} LS \left( p^{\prime} \right) / |P^{\prime}+|$, where $LS(\cdot)$ is defined as

\begin{footnotesize}
    \begin{equation}\label{eq:8}
LS(p^{\prime}) = \left\{
\begin{aligned}
+ \underset{q \in  \partial M} {\inf}\Vert p^{\prime} - q \Vert{_2}   ,   
    p^{\prime} \in  M_{in}\\
0,  p^{\prime} \in  \partial M\\
+ 2 \underset{q \in  \partial M} {\inf}\Vert p^{\prime} - q  \Vert{_2},      
    p^{\prime} \in  M_{out},\\
\end{aligned}
\right.
\end{equation}
\end{footnotesize}
where $\partial M$ is the zero level set. $M_{in}$ and $M_{out}$ represent the inside area and outside area for $M$. We also use the appearance loss $\mathcal{L}_a$ (\ie L1 loss), line and shape penalty $\mathcal{L}_p$ \cite{nie2022deep,liao2023recrecnet} on the mesh formed by $P^{\prime}$ to prevent excessive rectification, and the gradient loss $\mathcal{L}_g$ to align local textures. Finally, the all loss $\mathcal{L}_{DM}$ can be expressed by 

\begin{footnotesize}
    \begin{equation}\label{eq:9}
        \mathcal{L}_{DM} = \sum_{j=0}^1 \gamma^j (\mathcal{L}_a^j + \alpha_1\mathcal{L}_b^j +\alpha_2\mathcal{L}_p^j + \alpha_3\mathcal{L}_g^j),
    \end{equation}
\end{footnotesize}
where $j=0$ for $X_1^i$ and $j=1$ for $X_D^i$. $\gamma, \alpha_1,\alpha_2,\alpha_3$ are loss weights. Note that we only apply $\mathcal{L}_a$ for $\mathcal{T}_1, \mathcal{T}_4$ since these two tasks have no border changes during rectification. For restoration module, we only use appearance loss and perceptual loss like some super-resolution works \cite{Wang2018ESRGANES, johnson2016perceptual}.

\subsection{Sparse Mixture-of-Experts}
After obtaining four-by-one, we can accommodate four-in-one model. 
Sharing a same network is meaningful since it would be more convenient to further design model compression and inference acceleration algorithms for deployment in devices with limited computing resources. 
However, we still find it difficult to jointly train our UniRect on four task datasets since there are heavy \textbf{task competitions} and degradations observed in \cref{fig:degrad}a-d and \cref{tab:multi}. This could be attributed to the conflicts in the optimization direction resulting from the complex distortions associated with different tasks.
Therefore, we design a Spare Mixture-of-Experts (SMoEs) structure to aggregate our UniRect and circumvent task competitions:

\begin{footnotesize}
\begin{equation}
\label{eq:10}
SMoEs(X_0^i, M_0^i)=\sum_{j=1}^5 G(X_0^i)_j E_j(X_0^i, M_0^i), 
\end{equation}
\end{footnotesize}
where $G(\cdot)$ is a gating network and $E_j$ is the $j$-th expert network (\ie our UniRect). This gating network can apportion the specific input task among $E_j$ with a $5$-dimension vector weights. Specifically, to reduce computation, $G$ with network weights $W_G$ has a additional top-$k$ operator, enforcing the invalid value to be zero \cite{shazeer2017outrageously}, \ie

\begin{footnotesize}
\begin{equation}
\label{eq:11}
\begin{split}
    G(X_0^i)=\text{SoftMax}\left( \text{Top k}\left(X_0^i\cdot W_G,k \right ) \right ),  \\
\text{Topk}(x,k)_j = \left\{
\begin{aligned}
    &x_j, \text{if } x_j \text{ in Topk elements} \\ 
    &-\infty, \text{otherwise}.
\end{aligned}
\right.
\end{split}
\end{equation}
\end{footnotesize}

We can train the gating network along with the rest of the model. Gradient flow can also back-propagate through the gating network it inputs. In our experiments, we choose $k=1$, which can achieve a better balance between prediction and computation.

\section{Experiments}
\label{sec:experiment}

\begin{table*}[tp]
\centering
\scalebox{0.85}{
\begin{tabular}{cccccc}
\toprule[1pt]
Tasks                                                   & Methods  &  \makecell[c]{Line-\\ACC $\uparrow$} & \makecell[c]{Shape-\\ACC $\uparrow$} & \makecell[c]{LineACC-\\LLM $\uparrow$} & \makecell[c]{ShapeACC-\\LLM $\uparrow$} \\ \midrule
\multirow{4}{*}{\makecell[c]{Portrait correction \\ $\mathcal{T}_1$}}              &  \citet{shih2019distortion}            & 66.143    & 97.253      & -     & -  \\

    & \citet{tan2021practical}  & 66.784    & 97.490    & -     & -     \\
    & \citet{zhu2022semi}       & \textbf{66.825}     & \textbf{97.491}      & 7.082     & 8.082           \\
\rowcolor{Gray}\cellcolor{white}
   & UniRect (Ours)        & 66.523   & 97.454     & \textbf{7.168}     &  \textbf{8.152}     \\ \midrule
    
Tasks &   Methods              & PSNR $\uparrow$ & SSIM $\uparrow$  & FID $\downarrow$  & LPIPS $\downarrow$ \\ \midrule

\multirow{3}{*}{\makecell[c]{Rectified wide- \\ angle image 
 \\ rectangling $\mathcal{T}_2$}} 
& ROP \cite{liao2021towards} & 13.90     & 0.3516      & -     &  -       \\
& RecRecNet \cite{liao2023recrecnet}              & 18.68     & 0.5450      & \textbf{19.01}     & \textbf{0.1136}      \\
 &  \cellcolor{Gray} UniRect (Ours)         & \cellcolor{Gray} \textbf{19.90} & \cellcolor{Gray} \textbf{0.5721} & \cellcolor{Gray} 27.02 & \cellcolor{Gray} 0.1245      \\ \midrule
    
\multirow{4}{*}{\makecell[c]{Stitched image \\ rectangling $\mathcal{T}_3$}}       & \citet{he2013rectangling}            & 14.70     & 0.3775      & 38.19     & 0.2846      \\
    & \citet{nie2022deep}           & 21.28 & 0.7141 & 21.77 & 0.1557     \\
    & MOWA \cite{liao2024mowa} & 20.42 & 0.6307 & -     & -    \\
\rowcolor{Gray}\cellcolor{white}    & UniRect (Ours)         & \textbf{25.10}  & \textbf{0.7526} & \textbf{19.59} & \textbf{0.1120}   \\ \midrule
    
\multirow{4}{*}{\makecell[c]{Rotation correction \\ $\mathcal{T}_4$}}              & \citet{he2013content} & 21.69     & 0.6460      & 8.51     & 0.2120    \\
    & DRC \cite{nie2023deep}                   & 21.02 & 0.6280  & 7.12  & 0.2050  \\
    & CoupledTPS \cite{nie2024semisupervised}            & 22.29 & 0.6790  & 7.90   & 0.1970  \\
 \rowcolor{Gray}\cellcolor{white}   & UniRect (Ours)         & \textbf{23.16} & \textbf{0.7179}  & \textbf{6.55}  & \textbf{0.0873} \\ \bottomrule
\end{tabular}
}
\caption{The quantitative results of the proposed UniRect and other solutions of different tasks on five datasets. In the test dataset of $\mathcal{T}_1$,  no image ground-truth is provided. Metrics such as PSNR, SSIM, FID, and LPIPS are not applicable to this task. Due to these limitations, LLM-based metrics are also introduced. The best performance is in \textbf {bold}.
}
\label{tab:comparison}
\end{table*}

\begin{table*}[t]
\centering
\scalebox{0.85}{
\begin{tabular}{cccccc}
\toprule[1pt]
Strategy                         & $\mathcal{T}_1$ (ShapeACC$\uparrow$) & $\mathcal{T}_2$ (PSNR$\uparrow$)  & $\mathcal{T}_3$ (PSNR$\uparrow$) & $\mathcal{T}_4$ (PSNR$\uparrow$) & Parms \\ \midrule
ML                  & 97.223                                      & 13.07                                                                      & 15.74                                   & 21.73 &     357.9M                              \\
SL (1-4-3-2) & 97.401                                      & 15.23                                                                      & 16.94                                   & 21.51       &     357.9M                               \\
SL (2-3-4-1) & 97.375                                      & 15.72                                                                       & 16.59                                   & 20.73           &     357.9M                           \\
SL (3-2-4-1) & \textbf{97.409}                                      & 17.54                                                                      & 23.55                                   & 15.10       &     357.9M                                \\
SL (4-1-3-2) & 97.385                                      & 15.00                                                                         & 16.81                                   & 22.93             &     357.9M                         \\
\rowcolor{Gray} UniRect (SMoEs)                           & 97.390                                      & \textbf{19.90}                                                                       & \textbf{25.07}                                    & \textbf{23.16}             &     369.6M                 \\ \bottomrule[1pt]
\end{tabular}
}
\caption{Studies on learning strategy. For sequential learning, the number means the sequence in which training tasks are added. We have trained model using a longer 250 epochs for mixed learning and all sequential learning strategies. 'ML' denotes mixed learning and 'SL' represents sequential learning. The number after the 'SL' is  task order.}
\label{tab:multi}
\vspace{-1em}
\end{table*}
\subsection{Experimental Settings}
\paragraph{Datasets.} 
We conduct experiments on public representative benchmarks including portrait correction dataset \cite{tan2021practical}, rectified wide-angle image rectangling dataset \cite{liao2023recrecnet}, stitched image rectangling dataset \cite{nie2022deep}, and rotation correction dataset \cite{nie2023deep}. 

\paragraph{Implementation Details.} Our networks are based on the PyTorch framework with four NVIDIA Tesla V100 GPUs. For single-task learning, it takes one to three days to train our model on each dataset with a batch size of $4$ and 200 epochs. For multi-task learning, it takes about $7$ days to train on four datasets jointly with a bacth size of $10$ and 200 epochs. 
We adopt the poly policy of the learning rate, the factor of which is $0.96$, and the initial learning rate is $1e-5$ for rotation correction and $1e-4$ for the rest tasks. Adam \cite{kingma2014adam} optimizer is selected and the weight decay is $1e-5$.
For RP-TPS, the number of control points is set to $12 \times 10$. For \cref{eq:9}, we set $\gamma, \alpha_1, \alpha_2, \alpha_3$ as $0.9, 1e-2, 1.0, 1e-2$ respectively. For SMoEs, we utilize a simple ResNet18 \cite{he2016deep} as the backbone of $G$. For computation complexity, the amounts of parameters (Params), floating point operations (FLOPs), and frames per second (FPS) are 357.9M, 62.98G, and 35.8 respectively. More analysis for computations can be seen in the supplementary files.

\subsection{Comparison with State-of-the-Art Methods}
\label{sec:sota}
We adopt the evaluation setting from the previous studies, utilizing the PSNR, SSIM \cite{Wang2004ImageQA}, FID \cite{heusel2017gans} and LPIPS \cite{Zhang2018TheUE} to assess these methods for $\mathcal{T}_2, \mathcal{T}_3, \mathcal{T}_4$. The first two are objective metrics while the following are image perceptual metrics. For portrait correction, Line Straightness Metric (LineACC) and Shape Congruence Metric (ShapeACC) are suggested in \cite{tan2021practical} to measure the correction results of salient lines and faces.
Besides, many studies \cite{zhang2023gpt, wu2024comprehensive, xie2024modification} have proposed exploiting MLLMs to assess the quality of unlabeled images, which has led to remarkable progress in the field.
Owing to the limitations \footnote{In the test set of this dataset, only the coordinates of the corresponding points in the reference and input distorted images are provided for metric calculation. Nevertheless, in our approach, because of the characteristics of the TPS model (irreversibility), it is arduous to acquire the precise coordinates of the key points on the distorted image after rectification, as depicted in \cref {fig:arc}f. Hence, we resort to an estimation method, resulting in a substantial reduction in the metrics of us.} of these two evaluation metrics, we further utilize Multimodal Large Language Models (MLLMs)-based metrics, namely LineACC-LLM and ShapeACC-LLM. By capitalizing on the potent capabilities of MLLMs, our objective is to carry out an assessment of image quality in terms of line straightness and shape congruence. Further details about these metrics can be found in the supplementary files.

\begin{figure}[tp]
    \centering
    \includegraphics[width=\linewidth]{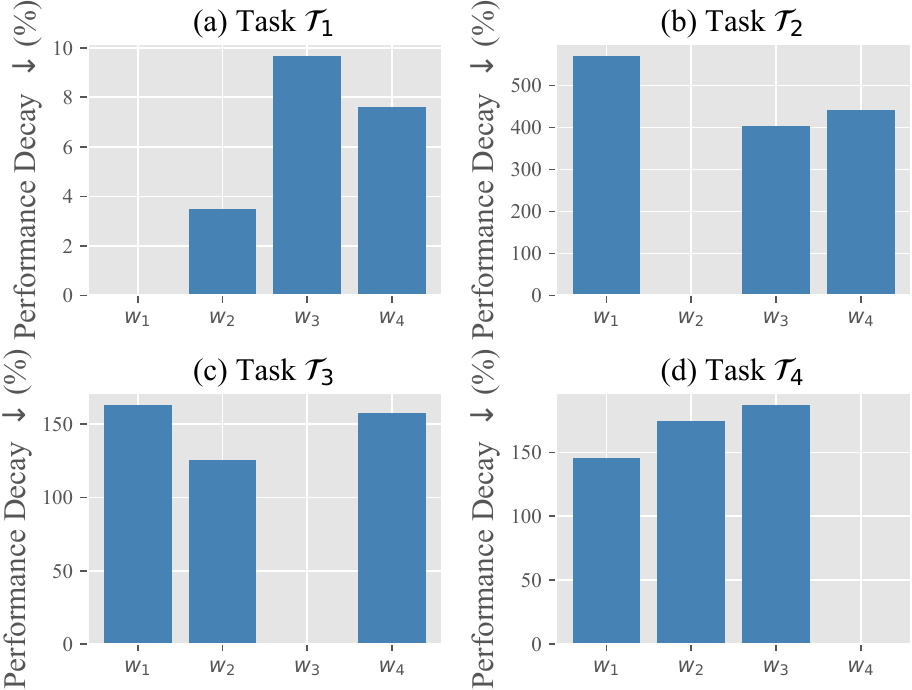}
    \caption{Cross-Task Degradation. Models trained on four tasks (different weights $w_i$) are evaluated for one task $\mathcal{T}_j$.  For $\mathcal{T}_1$, we use LineACC to quantify the performance, and FID is for remaining tasks.
    }
    \label{fig:degrad}
\end{figure}


Quantitative results are presented in \cref{tab:comparison}. 
Our proposed UniRect is compatible for four different types of tasks in a unified network structure and achieves promising results compared with methods for a single task. Specifically, we have made significant progress on rectified wide-angle image stitching, stitched image stitching, and rotation correction. 
For portrait correction, we also acquire satisfactory results. Moreover, our UniRect offers a novel potential technological path to solve the relevant low-level problems from the distortion rectification angle. Qualitative results are shown in \cref{fig:compare} for four tasks. More results and specific analysis can be seen in the supplementary files.

\subsection{Multi-Task Learning Strategy}
\noindent\textbf{Task Competitions.}
We investigate task competitions and showcase the efficacy of our proposed SMoEs strategy when applied to unified networks, \ie our UniRect, and the experiments are presented in \cref{fig:degrad}a-d and \cref{tab:multi}. As can be observed from the experiments, learning multiple tasks leads to a degradation in the performance of any individual task. In other words, directly using a model trained on other tasks can result in significant performance degradation, as different tasks often involve distinct types of distortions. We analyze this phenomenon in the supplementary files from the perspectives of distortion and data distribution.

\noindent\textbf{ML vs. SL vs. SMoEs.}
Mixed learning (ML) is a simple approach to multi-task learning, which involves training a model using all task datasets simultaneously. 
Still, we find learning multiple tasks does not contribute to an overall improvement in performance. Another strategy is sequential learning (SL), aiming at providing good starting points for subsequent tasks. However, the first task in sequential learning always becomes a predominant task. Thus, we propose employing SMoEs to adaptively switch different tasks, which can mitigate performance degradation when handling multi-task learning. 
Our SMoEs are nearly close to the single task learning for an effective gating network. More details and analysis can be seen in the supplementary files.

\begin{table}[t]
\centering
\small
\begin{tabular}{ccccc}
\toprule[1pt]
Strategy                         & $\mathcal{T}_1$ ($\uparrow$) & $\mathcal{T}_2$ ($\uparrow$)  & $\mathcal{T}_3$ ($\uparrow$) & $\mathcal{T}_4$ ($\uparrow$) \\ \midrule
w/o prompts  & 96.324   & 19.84  &  23.82    & 23.15   \\
w prompts & \textbf{97.454 }   & \textbf{19.90}   &  \textbf{25.10 }   & \textbf{23.16 }  \\
\bottomrule[1pt]
\end{tabular}
\caption{Results of the model with and without prompts. $\mathcal{T}_1$ uses ShapeACC, while the remaining tasks uses PSNR. }
\label{tab:prompt}
\end{table}
\begin{table}[tp]
\centering
\small
\begin{tabular}{ccccccc}
\toprule[1pt]
\multirow{2}{*}{Task}  & \multicolumn{2}{c}{RP-TPS}  & \multicolumn{4}{c}{Metrics}      \\ \cmidrule{2-7} 
 & w/o           & w                          & PSNR $\uparrow$ & SSIM $\uparrow$ & FID $\downarrow$  & LPIPS $\downarrow$  \\ \hline
\multirow{2}{*}{$\mathcal{T}_5$}                     & $\checkmark$     &        &      21.56 & 0.6502 & 7.21 & 0.0972 \\
  &               & $\checkmark$          & \textbf{23.16} & \textbf{0.7179} & \textbf{6.55} & \textbf{0.0873} \\ 
                                     \bottomrule[1pt]
\end{tabular}
\caption{The ablation study on the influence of RP-TPS.}
\label{tab:num_block}
\end{table}

\begin{table}[tp]
\centering
\scalebox{0.75}{
\begin{tabular}{ccccccc}
\toprule[1pt]
Task   &  Method   & PSNR $\uparrow$  & SSIM $\uparrow$ & FID $\downarrow$ & LPIPS $\downarrow$ & Param $\downarrow$   \\ \midrule
\multirow{3}{*}{$\mathcal{T}_4$}  & CNN   & 24.12  & 0.7123 & 21.03 &  0.1250    & 508.9M    \\
 & Transformer  & 24.23 & 0.7226 &  21.27 &   0.1320           & 1.168G    \\
\rowcolor{Gray}\cellcolor{white}  & Mamba     & \textbf{25.10}     & \textbf{0.7526} & \textbf{19.59}    & \textbf{0.1120}            & \textbf{357.9M}    \\ \midrule
\multirow{3}{*}{$\mathcal{T}_5$}  & CNN   & 22.31  & 0.6796 & 7.20 &  0.0972    & 508.9M    \\
 & Transformer  & \textbf{23.23} & 0.6884 &  8.59 &   0.1546           & 1.168G    \\
\rowcolor{Gray}\cellcolor{white}  & Mamba     & 23.16     & \textbf{0.7179} & \textbf{6.55}    & \textbf{0.0873}            & \textbf{357.9M} \\
  \bottomrule[1pt]
\end{tabular}

}
\caption{The ablation study on the architectures of pure convolutions (CNN), Transformer, and Mamba.}
\label{tab:mamba}
\end{table}

\subsection{Ablation Studies}
We evaluate the components of UniRect on diverse tasks. More ablation experiments are in the supplementary files.

\begin{table}[tp]
\centering
\small
\begin{tabular}{cccccc}
\toprule[1pt]
Task   &  Method   & PSNR $\uparrow$  & SSIM $\uparrow$ & FID $\downarrow$ & LPIPS $\downarrow$   \\ \midrule

\multirow{2}{*}{$\mathcal{T}_2$}  & w/o RM   & \textbf{19.99}  & \textbf{0.5791} & 28.69 &  0.1296     \\

  & \cellcolor{Gray} w RM     & \cellcolor{Gray} 19.90     & \cellcolor{Gray} 0.5721 &  \cellcolor{Gray}\textbf{12.51}    & \cellcolor{Gray} \textbf{0.1245}      \\ \midrule
\multirow{2}{*}{$\mathcal{T}_4$}  & w/o RM   & 22.84  & 0.6901 & 7.88 &  0.1560        \\
 & \cellcolor{Gray} w RM     & \cellcolor{Gray} \textbf{23.16}     & \cellcolor{Gray} \textbf{0.7179} & \cellcolor{Gray} \textbf{6.55}    & \cellcolor{Gray} \textbf{0.0873}      \\ 
  
  \bottomrule[1pt]
\end{tabular}
\caption{Study on the role of our restoration module (RM).}
\label{tab:RM}
\end{table}

\begin{figure}[htp]
    \centering
    \includegraphics[width=\linewidth]{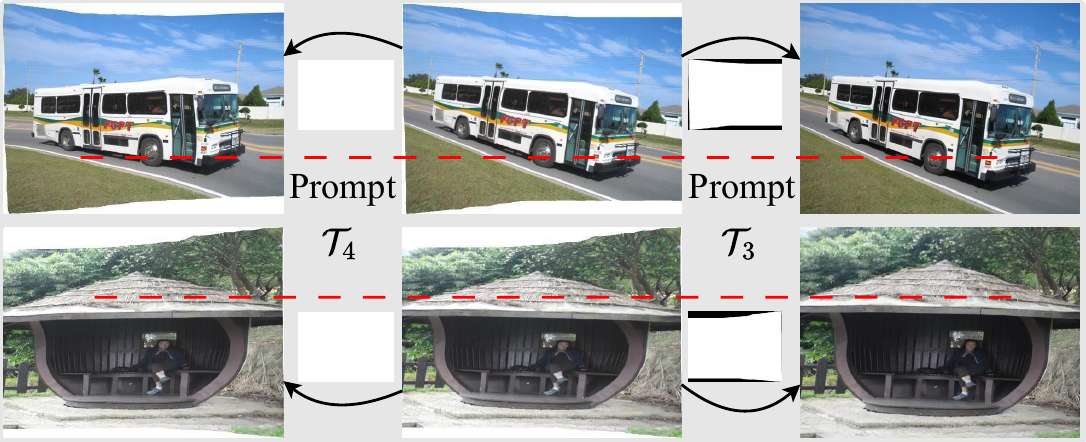}
    \caption{Results of the same input under $\mathcal{T}_3$ and $\mathcal{T}_4$ prompts. 
    }    \label{fig:general}
\end{figure}


\noindent\textbf{Prompts Designs.} 
According to our experiments in \cref{tab:prompt}, without prompt, the results of task $\mathcal{T}_1$ and task $\mathcal{T}_3$ were significantly reduced. They have a slight impact on task $\mathcal{T}_2$ and no impact on task $\mathcal{T}_4$. Therefore, prompts have a significant effect on the network performance. \textbf{Why not Task IDs?} The original datasets for $\mathcal{T}_1$ and $\mathcal{T}_3$ already have masks: one indicating the face location and the other indicating the effective area. Thus, prompts are designed  for other tasks to align input channels. 
Visual prompt is of paramount significance for our network as it takes part in the calculation of the loss function and can also be utilized to determine and even control the rectification position. However, task-ID is hard to indicate the location and enhance the performance.


\noindent\textbf{RS-TPS.}
We tested the validity of the RP-TPS model as seen in \cref{tab:num_block}. With the help of it, distortion can be estimated more accurately, leading to a better performance. We also discover that recursively repeating $\mathcal{G}$ and $\mathcal{C}$ resulted in no obvious improvement and introduce more parameters due to the matrix inversion operator in the recursive manner. 


\noindent\textbf{Comparisons with CNN/Transformer.}
Mamba has shown impressive performance on many vision tasks due to its long-range modeling capability and efficient computing complexity. Thus, we also try to explore its ability of modeling  geometric information in this paper. The ablation results are shown in \cref{tab:mamba}. Compared with CNN and Transformer, Mamba is a strong choice for balancing both accuracy and computational efficiency. 

\noindent\textbf{Restoration Module.}
The contribution of our restoration module differs in different tasks (\cref{tab:RM}).  
For $\mathcal{T}_4$, our restoration module can greatly improve the final image quality and compensate for the sampling degradation in the deformation module. However, for $\mathcal{T}_2$, our deformation module dominates the improvement. Actually, this module is extremely important, since the restoration module cannot learn accurate mappings if the former module gives wrong and chaotic pixel locations. 

\subsection{More Applications}
We found that the introduction of prompts enables the implementation of some interesting applications. \cref{fig:general} represents a scenario in which two forms of distortion are concurrently present within a single image. By providing different prompts for the same input under different tasks, our model can accomplish the corresponding tasks and achieve controllable rectification of images (designated tasks and areas).
Furthermore, the cross-domain results in real-world scenes can be found in the supplementary files. This adequately demonstrates the generalization capability of our model.

\section{Conclusion}
\label{sec:conclusion}

In this paper, we study four tasks from a novel consistent distortion perspective. We build a general distortion model to unify different photography distortion problem, and propose a unified rectification framework based on RP-TPS to handle these tasks in the same structure designs. Moreover, we introduce sparse MoEs to address task competitions of multi-task learning. 
It is also intriguing to consider another complex scenario in which a model can apply multiple rectifications to the same sample simultaneously. Since all existing datasets contain only one type of distortion, conducting an experimental verification at this stage is quite challenging. We look forward to exploring this highly interesting issue in our subsequent work.

\section*{Acknowledgments}
This work was supported in part by the National Natural Science Foundation of China under Grant 62475006 and 62125102, in part by the National Key Research and Development Program of China under Grant 2022ZD0160401.

\bibliography{aaai2026}


\clearpage
\appendix
\section{Supplementary Materials for UniRect} 

\subsection{Overview}
In this file, we provide the following supplementary studies:
\begin{itemize}
\item Analysis for our general distortion model.
\item More ablation studies and analysis of our framework and loss function designs.
\item More results and analysis for our SMoEs.
\item Comparisons of the parameters, inference time, and FLOPs of different methods.
\item MLLMs-based metrics for portrait correction. 
\item More qualitative results of the baselines and our method.
\item Cross-domain results under real-world scenes.
\item Comparisons with other methods and limitations.

\end{itemize}

Note that \blux{blue} color references represent the Figures, Equations or Tables in the main paper.

\subsection{Analysis for General Distortion Model}

In this section, we discuss the various tasks under our proposed general distortion model.
The general distortion model incorporates the Kannala-Brandt model \cite{kannala2006generic}, typically used to approximate wide-angle lenses, and the Brown-Conrady model \cite{weng1992camera}, employed for radial distortion modeling. 
Consequently, we can employ this proposed distortion model to simulate real image distortion in real-world smartphone camera systems. As shown in \blux{Eq. (3)}, This model can be simplified as

\begin{small}
\begin{equation}\label{eq:s0}
\left[ {\begin{array}{*{20}{c}}
{{x_d}}\\
{{y_d}}
\end{array}} \right] = \frac{r_d}{r}\cdot \mathcal{R}_{\alpha} \left[ {\begin{array}{*{20}{c}}
x\\
y
\end{array}} \right] + T_0, N=1,2,3,\cdots,
\end{equation}
\end{small}
where $r_d=\sum\limits_{j = 1}^N \left( {k_j{\theta ^{2j - 1}}}  + {k_j^{\prime}{r^{2j - 1}}}  \right)$ and it denotes pixel  distorted location. 
\begin{small}
$\mathcal{R}_{\alpha}=\left[{\begin{array}{*{20}{c}}
{\cos \alpha }&{ - \sin \alpha }\\
{\sin \alpha }&{\cos \alpha }
\end{array}} \right]$ \end{small}
represents the pixel rotation matrix. $T_0$ is the translation matrix with eccentric shifting determined by the borders.

1) For portrait distortion in $\mathcal{T}^{-1}_1$, let $k_j^{\prime}=0, \alpha=0, T_0=[0, 0]^T$, and we can get 
\begin{small}
\begin{equation}\label{eq:s1}
\left[ {\begin{array}{*{20}{c}}
{{x_d}}\\
{{y_d}}
\end{array}} \right] = \frac{1}{r} \left({\sum\limits_{j = 1}^N {k_j{\theta ^{2j - 1}}} } \right) \left[ {\begin{array}{*{20}{c}}
x\\
y
\end{array}} \right], N=1,2,3,\cdots.
\end{equation}
\end{small}

2) For wide-angle image rectified distortion in $\mathcal{T}^{-1}_2$, let $k_j=0, \alpha=0, T_0=[0, 0]^T$, and we can get 
\begin{small}
\begin{equation}\label{eq:s2}
\left[ {\begin{array}{*{20}{c}}
{{x_d}}\\
{{y_d}}
\end{array}} \right] = \frac{1}{r} \left({\sum\limits_{j = 1}^N {k_j^{\prime}{r ^{2j - 1}}} } \right) \left[ {\begin{array}{*{20}{c}}
x\\
y
\end{array}} \right], N=1,2,3,\cdots.
\end{equation}
\end{small}

3) For stitched distortion in $\mathcal{T}^{-1}_3$, let $k_j=0,\alpha=0$, and we can get 
\begin{small}
\begin{equation}\label{eq:s4}
\left[ {\begin{array}{*{20}{c}}
{{x_d}}\\
{{y_d}}
\end{array}} \right] = \frac{1}{r} \left( {\sum\limits_{j = 1}^N {k_j^{\prime}{{r}^{2j - 1}}} } \right) \left[ {\begin{array}{*{20}{c}}
x\\
y
\end{array}} \right] + {T}_0, N=1,2,3,\cdots.
\end{equation}
\end{small}

\begin{figure}[tp]
    \centering
    \includegraphics[width=0.75\linewidth]{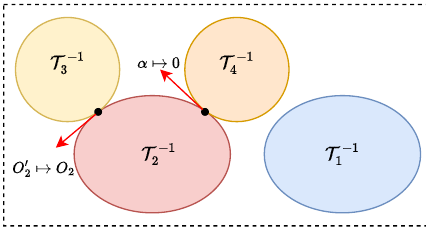}
    \caption{Underlying relationship map for four distortions. Red arrows denote the conversion conditions between different distortions. This map is conducted by our general distortion model and experiments of cross-task degradation in \blux{Figure 4.a-d}.
}
    \label{fig:distortion}
\end{figure}

4) For rotation distortion in $\mathcal{T}^{-1}_3$, let $k_j=0, T_0=[0, 0]^T$, and we can gain
\begin{footnotesize}
\begin{equation}\label{eq:s5}
\left[ {\begin{array}{*{20}{c}}
{{x_d}}\\
{{y_d}}
\end{array}} \right] = \frac{1}{r} \left( {\sum\limits_{j = 1}^N {k_j^{\prime}{{r}^{2j - 1}}} } \right) 
\begin{bmatrix}
{\cos \alpha } &  { - \sin \alpha }\\
{\sin \alpha } &  {\cos \alpha }
\end{bmatrix}
\left[ {\begin{array}{*{20}{c}}
x\\
y
\end{array}} \right], N=1,2,3,\cdots.
\end{equation}
\end{footnotesize}

Upon comparing various specific distortion models, the wide-angle image rectified distortion ($\mathcal{T}^{-1}_2$) can be subsumed under stitched distortion ($\mathcal{T}^{-1}_3$). In $\mathcal{T}^{-1}_3$, an additional eccentric displacements related to the boundaries are introduced, and the symmetric center of boundaries in $\mathcal{T}^{-1}_2$ is $O_2$, \ie $O_2^{\prime} \mapsto O_2$. Also, the wide-angle image rectified distortion ($\mathcal{T}^{-1}_2$) can be subsumed under rotation distortion ($\mathcal{T}^{-1}_4$) so long as $\alpha \mapsto 0$. The underlying relationship map among these distortions is shown in \cref{fig:distortion}, which can explain the performance degradation of \blux{Figure 4.a-d} to some extent. For example, $\mathcal{T}^{-1}_3$ and $\mathcal{T}^{-1}_2$ intersect at a specific point in \cref{fig:distortion} and the model trained on $\mathcal{T}_2$ (\ie $w_2$) shows the least performance degradation for $\mathcal{T}_3$ (see \blux{Figure 4.c}). \blux{Figure 4.d} also follows a similar pattern but may deviate due to the inherent data distribution as seen in \cref{fig:tSNE_input}. Images from $\mathcal{T}_1$ and $\mathcal{T}_4$ are extremely close. Therefore, the model trained on $\mathcal{T}_1$ obtains the least performance degradation for $\mathcal{T}_4$  (see \blux{Figure 4.d}).

The t-SNE visualization of the input data is presented in \cref{fig:tSNE_input}. Some tasks have data distributions that are markedly different from others. If a single model is trained on the combined data, it may be biased towards the dominant task or data distribution, leading to suboptimal performance on the less dominant tasks. We deem this an important reason, which is why we employ MoE.

\begin{figure}[tp]
    \centering
    \includegraphics[width=0.7\linewidth]{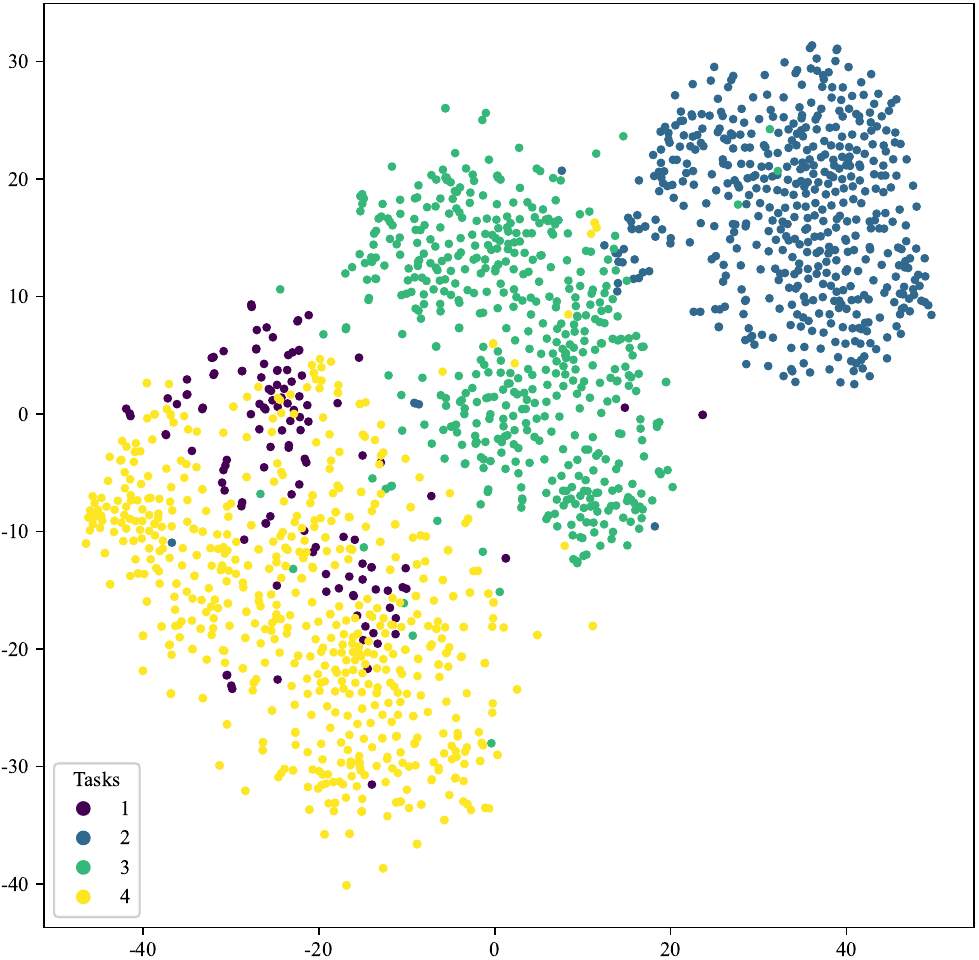}
    \caption{t-SNE visualization of input data. Images from $\mathcal{T}_1$ and $\mathcal{T}_4$ have tangled together. Zoom in for best view. }
    \label{fig:tSNE_input}
\end{figure}

\subsection{Ablation Experiments and Network Design Analysis}
\label{sec:abla}
In this section, we provide enough ablation experiments on the loss functions and the control point number. Besides, we also give some analyses about our network designs.

\subsubsection{The Role of DM}
DM focuses on \emph{geometric rectification}. If only the RM is used, coordinates of the pixels remain largely unaltered. The blank regions are misconstrued as valid information. For example, results in $\mathcal{T}_3$ will be produced with wrong representations, as seen in \cref{fig:RM}.

\subsubsection{Influence of Loss Functions.} 
\label{sec:app_loss}
We first give visualization (see \cref{fig:loss}) and detailed explanations about the loss for our deformation module. 
As introduced in the main paper, it mainly includes L1 loss $\mathcal{L}_a$, boundary loss $\mathcal{L}_{b}$, line and shape penalty $\mathcal{L}_{p}$, and gradient loss $\mathcal{L}_{g}$. For boundary loss $\mathcal{L}_{b}$ as seen in \cref{fig:loss}a, the purpose of this loss function is to make the outermost control points as close as possible to the real boundary of the input image, thereby obtaining better results. For line and shape penalty $\mathcal{L}_{p}$, it has line penalty $\mathcal{L}_{pl}$, and shape penalty $\mathcal{L}_{ps}$. 

For experiment 1 to 5 in \cref{tab:ablation_loss}, we explore the influence of different loss functions on stitched image rectangling dataset. $\mathcal{L}_b$ focuses on the boundaries while $\mathcal{L}_p$ and $\mathcal{L}_g$ can significantly improve performance. We also discover that the PSNR metric in \cref{tab:ablation_loss} is improved from 21.86dB to 25.08dB using the restoration module but the SSIM metric gains limited improvements. This problem is mainly due to the intrinsic defect (inconsistent hue) of the stitched image rectangling dataset. SSIM usually focuses on the image textures and structures, making it less affected by hue inconsistencies. 

\begin{figure}[tp]
    \centering
    \includegraphics[width=0.8\linewidth]{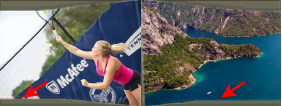}
    \caption{Results of using RM only.}
    \label{fig:RM}
\end{figure}

\begin{figure}[tp]
    \centering
    \includegraphics[width=1\linewidth]{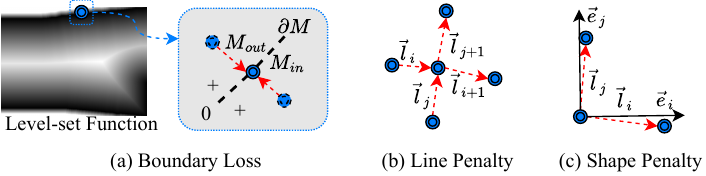}
    \caption{Visualization of our loss functions. $\mathcal{L}_{DM}$ includes (a) boundary loss $\mathcal{L}_{b}$, (b) line penalty $\mathcal{L}_{pl}$, and shape penalty $\mathcal{L}_{ps}$. Zoom in for best view. }
    \label{fig:loss}
\end{figure}

\begin{table}[!t]
\centering
\scalebox{0.8}{
\begin{tabular}{ccccccccc}
\toprule[1pt]
\multirow{2}{*}{No.} & \multicolumn{4}{c}{Loss Function}                                                                                                     & \multicolumn{2}{c}{Restortaion Module} & \multicolumn{2}{c}{Metric} \\ \cmidrule{2-9} 
                     & $\mathcal{L}_a$ & $\mathcal{L}_b$ & $\mathcal{L}_p$ & $\mathcal{L}_g$ & w/o                 & w                & PSNR $\uparrow$       & SSIM $\uparrow$        \\ \midrule
1                    & $\checkmark$                               &                                 &                                 &                                 & $\checkmark$                   &                  & 21.74       & 0.7297       \\
2                    & $\checkmark$                               & $\checkmark$                               &                                 &                                 & $\checkmark$                   &                  & 21.69       & 0.7257       \\
3                    & $\checkmark$                               &                                 & $\checkmark$                               &                                 & $\checkmark$                   &                  & 21.85       & 0.7381       \\
4                    & $\checkmark$                               & $\checkmark$                               & $\checkmark$                               &                                 & $\checkmark$                   &                  & 21.72       & 0.7278       \\
5                    & $\checkmark$                               & $\checkmark$                               & $\checkmark$                               & $\checkmark$                               & $\checkmark$                   &                  & 21.86       & 0.7386       \\
6                    & $\checkmark$                               & $\checkmark$                               & $\checkmark$                               & $\checkmark$                               &                     & $\checkmark$                & \textbf{25.08}       & \textbf{0.7528}       \\ \bottomrule[1pt]
\end{tabular}
}
\caption{The ablation study on loss functions and restoration module for $\mathcal{T}_3$.}
\label{tab:ablation_loss}
\end{table}
\begin{table}[!ht]
\centering
\scalebox{0.85}{
\begin{tabular}{ccccc} 
\toprule[1pt]
Control point number $N_c$ & PSNR $\uparrow$ & SSIM $\uparrow$  & FID $\downarrow$  & LPIPS $\downarrow$ \\ \midrule
7$\times$5                  & 21.12 & 0.6907 & 22.22 & 0.1554 \\
9$\times$7                  & 21.74 & 0.7313 & 21.18 & 0.1475 \\
10$\times$8                 & 21.86 & 0.7386 & 20.55 & 0.1462 \\
11$\times$9                 & 21.83 & 0.7356 & 20.62 & 0.1463 \\
12$\times$10                & 21.87 & 0.7376 & \textbf{20.40}  & 0.1457 \\
14$\times$12                & \textbf{21.90}  & \textbf{0.7398} & 20.54 & 0.1452 \\
15$\times$13 & 21.90 & 0.7388 & 20.58 & \textbf{0.1448}\\
16$\times$14                & 21.77 & 0.7312 & {20.66} & 0.1470  \\ 
\bottomrule[1pt]
\end{tabular} 
}
\caption{The influence of control point number $N_c$. For $N_c=12\times 10$, $12$ means the point number along the width axis and $10$ means the point number along the height axis. }
\label{tab:control}
\end{table}

\subsubsection{Impact of Control Point Number}
\label{sec:app_control}
In this section, we do an ablation study of the number of control points in the deformation module.
As demonstrated in \cref{tab:control}, the performance of the output noticeably improves with an increase in the number of control points. 
However, a performance bottleneck is encountered when the number $N_c$ exceeds 14$\times$12, which is associated with a significant increase in computational complexity. 
Consequently, we establish the number $N_c$ as $12 \times 10$ taking into account both performance and computational complexity. 
This performance bottleneck could be the result of inadequate optimization during training, as the flexibility increases with the number $N_c$.

\subsubsection{The Role of Visual Prompts}
\label{sec:app_prompt}
Our visual prompt can serve several purposes: (1) Indicate the location of an image's irregular boundary or direct the network to a critical area that needs correction. (2) Change the input data distribution, as shown in \cref{fig:tSNE_prompt}. After introducing the prompt, the data distributions of different tasks can be decoupled to some extent. (3) Participate in the calculation of loss functions for certain tasks in \blux{Eq. (8)}.

\begin{figure}[htp]
    \centering
    \includegraphics[width=0.7\linewidth]{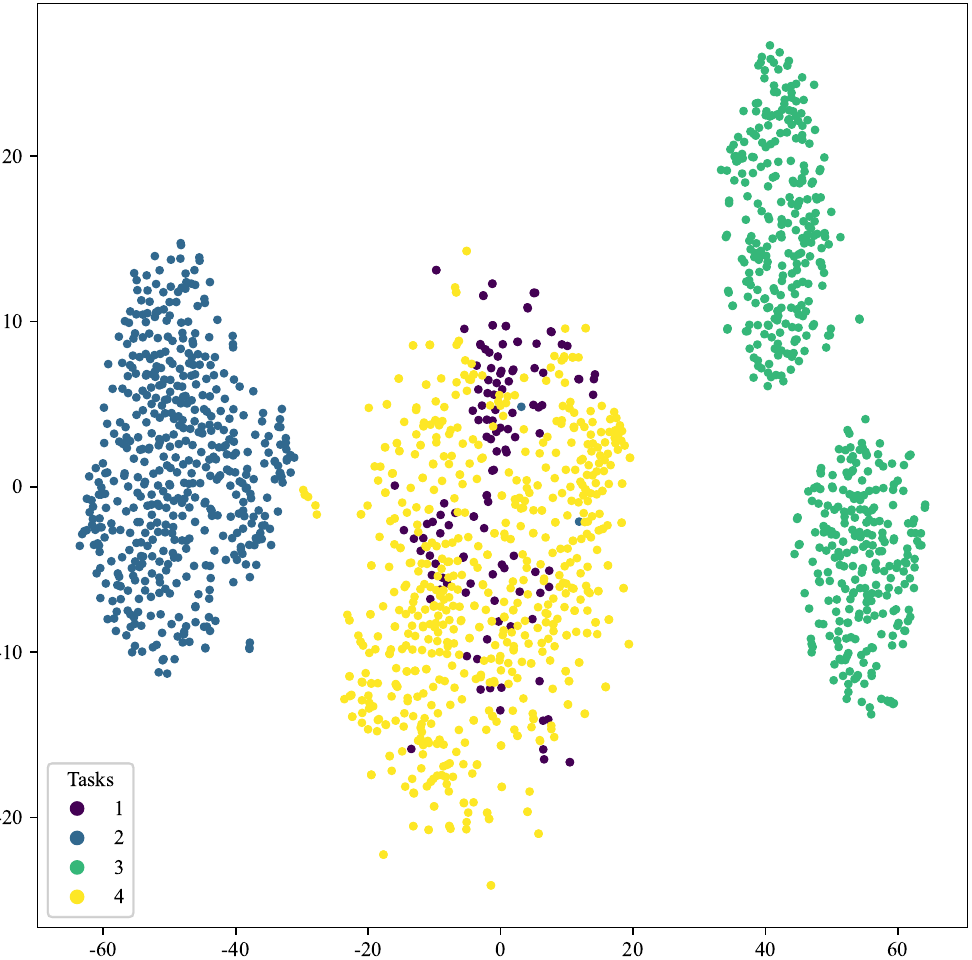}
    \caption{t-SNE visualization of input data with prompts. Images except $\mathcal{T}_1$ and $\mathcal{T}_4$ have been decoupled. Zoom in for best view.}
    \label{fig:tSNE_prompt}
\end{figure}

\subsubsection{The Role of Optical Flows}
Optical flow is a concept related to the analysis of the movement of objects in a sequence of images. We utilize this tool to explore the distortion process for all pixels and formulate each distortion for every task. This is a qualitative tool. 
The generation of optical flow via different methods is presented in \cref{fig:flow}. Taking the distortion $\mathcal{T}_2^{-1}$ as an example, we present the results of various optical flow estimation methods, such as FlowNet \cite{dosovitskiy2015flownet}, FlowNet2 \cite{ilg2017flownet}, PWC-Net \cite{sun2018pwc}, and RAFT \cite{teed2020raft}. It can be observed from the figure that the outcomes of different optical flow estimation methods are similar without altering the mathematical form of distortion derived in this paper.

\begin{figure}[tp]
    \centering
    \includegraphics[width=0.8\linewidth]{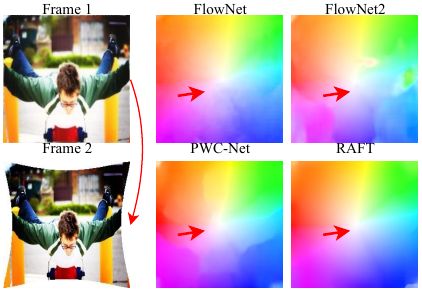}
    \vspace{-1em}
    \caption{Results of the optical flow by different methods. The flow is generated from groundtruth to input. Wide-angle image rectified distortion in $\mathcal{T}_2^{-1}$ is taken as an example. 
    }    \label{fig:flow}
    \vspace{-1em}
\end{figure}

\subsection{Dataset Details}
\noindent\textbf{Dataset Construction.} 
For portrait correction $\mathcal{T}_1$, the data was collected using five different ultra wide-angle smartphone cameras. Over ten people were photographed in several scenes \cite{tan2021practical}. The wide - angle portraits were manually corrected into distortion-free images.
For the rectified wide-angle image rectangling dataset \cite{liao2023recrecnet}, wide-angle images were first synthesized, and the ground-truth was obtained manually.
In the stitched image rectangling dataset \cite{nie2022deep} utilized for $\mathcal{T}_3$, the majority of the inputs are real stitching results, while the ground - truth is synthesized manually.
For rotation correction $\mathcal{T}_4$ \cite{nie2023deep}, the images are sourced from ImageNet \cite{deng2009imagenet} and processed manually.

\noindent\textbf{Data with Multiple Simultaneous Distortions.} 
Although some datasets are synthetically generated, synthesizing data with multiple simultaneous distortions in the same image is not straightforward. This is because numerous processes require manual calibration and human screening, especially for obtaining the ground-truth. This will be our next step of work.

\subsection{Analysis for Four-in-One}
\label{sec:moe}
\subsubsection{Routing Visualization for Four-in-One}
We give the routing visualization of our SMoEs in \cref{fig:router}.  For portrait correction $\mathcal{T}_1$, the gating network has difficulty distinguishing it from rotation correction $\mathcal{T}_4$. The gating network's performance was poor since $\mathcal{T}_1$ and $\mathcal{T}_4$ exhibit similar data distributions as seen in \cref{fig:tSNE_input}. 

\begin{figure}[htp]
    \centering
    \includegraphics[width=\linewidth]{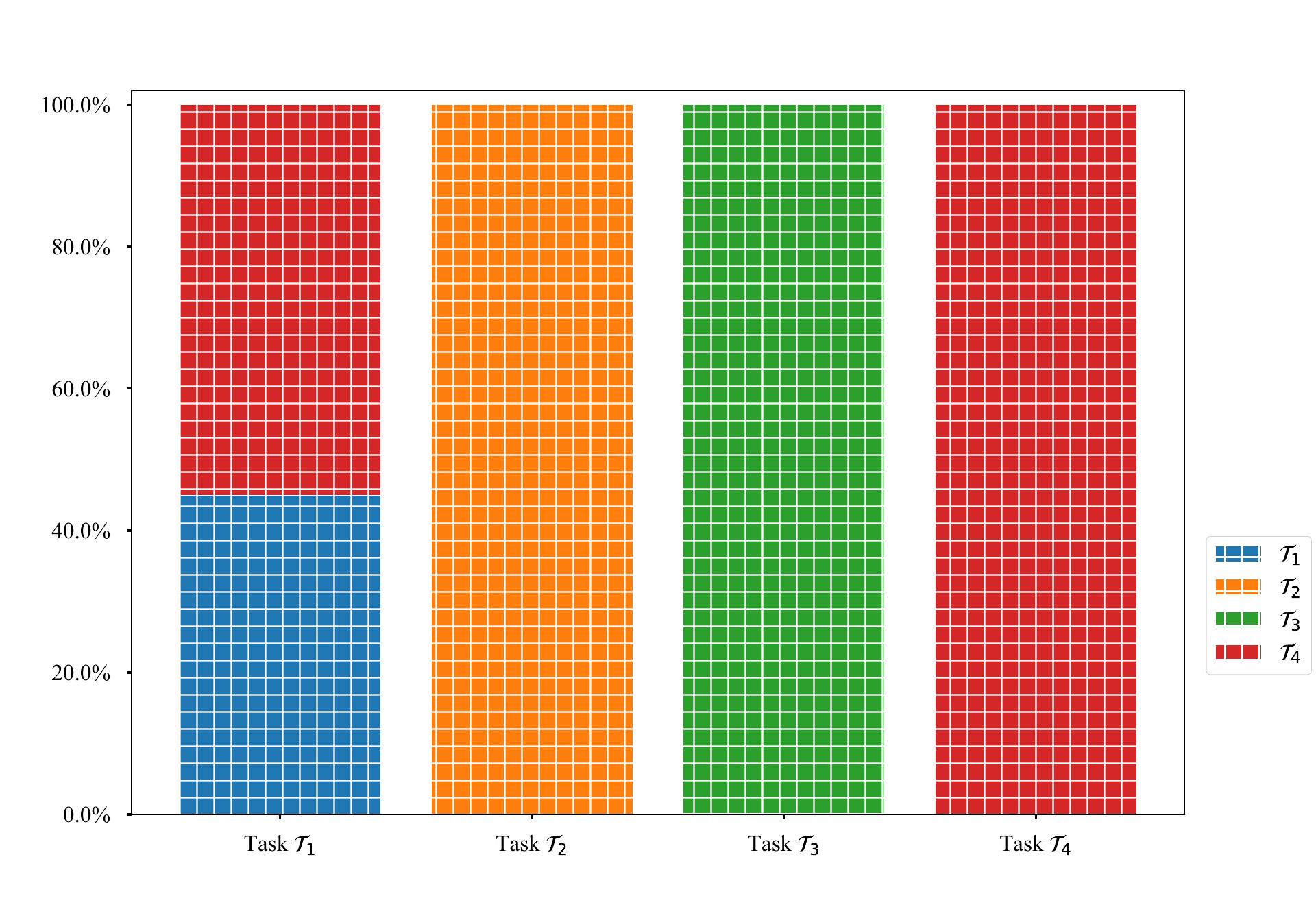}
    \caption{Routing visualization. }
    \label{fig:router}
\end{figure}

\subsubsection{Analysis for Multi-Task Learning Strategy}
In \blux{Table 2}, SL(3-2-4-1) performs better in Task $\mathcal{T}_1$ than the others. For $\mathcal{T}_1$, our strategy sometimes \emph{activated mismatched} expert models (see routing visualization in \cref{fig:router}), causing performance to drop. Compared to the base model, these strategies saw varying levels of decline, with SL(3-2-4-1) showing the least degradation.

\subsubsection{The Influence of $k$}
We choose $k = 1$ in \blux{Eq. (11)} since it is the \emph{simplest and lightest} option, yet still delivers performance close to that of a single model. Increasing $k$ leads to unstable training and higher computational cost, without improving performance.

\begin{table}[htp]
\centering
\scalebox{0.7}{
\begin{tabular}{lcccc}
\toprule[1pt]
Methods   & Task Type   & Param $\downarrow$  & FLOPs  $\downarrow$          & FPS $\uparrow$ \\ \midrule
Zhu \etal \cite{zhu2022semi}  & $\mathcal{T}_1$        & 8.79M  & -                & -    \\
RecRecNet \cite{liao2023recrecnet} & $\mathcal{T}_2$  & 62.70M & -                & -    \\
Nie \etal \cite{nie2022deep}   & $\mathcal{T}_3$       & 52.14M & -                & 20   \\
RecDiffusion \cite{zhou2024recdiffusion} & $\mathcal{T}_3$ & 800M   & \textgreater{}1T & -    \\
DRC  \cite{nie2023deep}   & $\mathcal{T}_4$     & -      & -                & 5    \\ \midrule
Ours  & $\mathcal{T}_1-\mathcal{T}_4$       & 357.9M & 62.98G           & 35.8 \\ \bottomrule[1pt]
\end{tabular}
}
\caption{Comparisons of the parameters, FLOPs, and inference time of different methods.}
\label{tab:compute}
\end{table}

\subsection{Comparisons of Computation Complexity}
\label{sec:compute}
Since few other methods have provided these comparisons officially,  we can only give  some estimated results as seen in \cref{tab:compute}. Some rectangling methods like RecDiffsuion   have two separated diffusion model in their rectangling process.  We think our method is not so cumbersome.  The main parameters of our model are concentrated in the restoration module ($>$300M).  The parameters of this structure can be reduced by setting a small block number. We even can handle four tasks without this module. Therefore, we think our model has potentials for limited-resources scene and edge-device.

\subsection{MLLMs-based Metrics for Portrait Correction}
\label{sec:MLLM}
Here, we introduce the details of two metrics (\ie LineACC-LLM and ShapeACC-LLM) based on Qwen-VL \cite{Qwen-VL} \footnote{Due to limited access, we do not employ popular MLLMs such as GPT-4o or Claude4-Sonnet.}. \cref{tab:LLM} serves as an example. It can be observed that MLLMs are capable of specifically scoring the Line Straightness and Shape Congruence of the input images in accordance with the instructions of the prompts. Due to the characteristics of LLM, these type of metrics are sensitive to prompts, and there is also a possibility of hallucinations. However, it can still objectively reflect the quality of the rectification results to a certain extent.

\begin{table*}[ht]
\centering
\begin{tabular}{p{6in}}
\toprule
\textbf{Input Image} \\

\multicolumn{1}{c}{
    \includegraphics[width=0.125\linewidth]{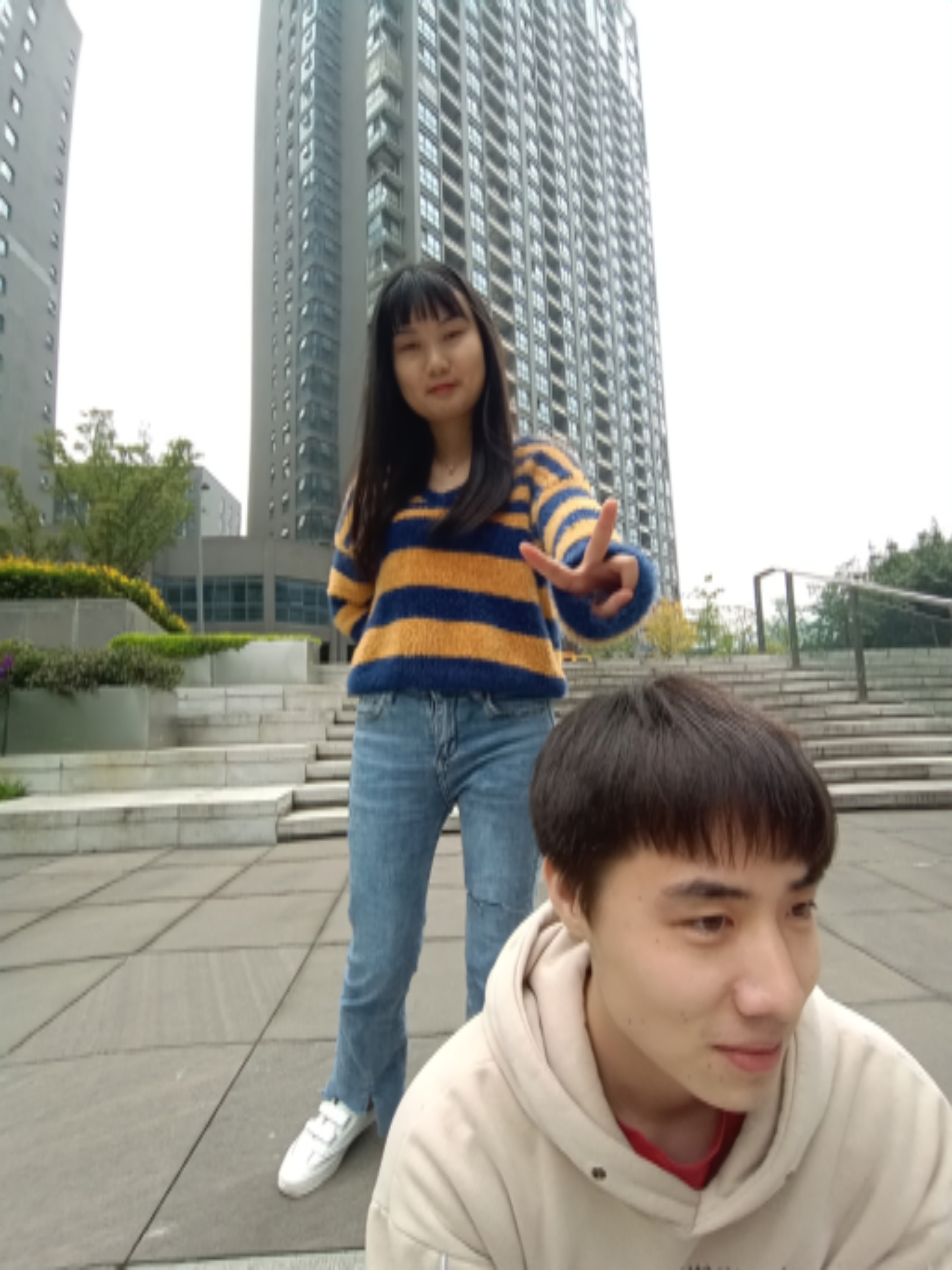}
} \\
\midrule
\textbf{Scoring Prompts} \\
\begin{promptbox}
\begin{multicols}{2}
\begin{lstlisting}[style=promptstyle]
IMPORTANT SCORING GUIDELINES:
- Scores of 8.0-10.0 should be EXTREMELY RARE and only given in exceptional cases
- Start from a neutral score of 5.0 for each metric
- Evaluate RELATIVE to professional portrait standards
- Be extremely critical and detail-oriented in your evaluation
- Any visible issues should result in immediate point deductions
- Multiple minor issues in the same category should compound the deductions
- When in doubt, always choose the lower score
- Be accurate to at least one decimal place.

Analyze this wide-angle portrait correction result focusing on the following aspects:

1. Line Straightness (1.0-10.0):
   - 1.0: Severe bending and distortion in all straight lines
   - 2.0-3.0: Major curvature in most straight lines
   - 4.0-5.0: Noticeable bending in several straight lines
   - 6.0-7.0: Minor curvature in few straight lines
   - 8.0-9.0: Very slight deviation in straight lines
   - 10.0: Perfect straightness in all lines
   - Examine each line against a virtual grid overlay
   - Check specifically:
      - Vertical lines (walls, doors, windows)
      - Horizontal lines (tables, shelves)
      - Background architectural elements
      - Frame edges and borders
   - Deduct points for:
      - 2 points: Each severely bent line
      - 1 point: Each noticeably curved line
      - 1 point: Each wavy edge or border
      - 0.5 points: Each subtle line deviation
      
2. Shape Congruence (1-10):
   - 1.0: Severe facial distortion with unnatural proportions
   - 2.0-3.0: Major perspective distortion affecting facial features
   - 4.0-5.0: Noticeable warping in facial structure
   - 6.0-7.0: Minor perspective artifacts in face shape
   - 8.0-9.0: Very slight unnatural facial proportions
   - 10.0: Perfect natural face shape without distortion
   - Compare against standard facial proportions
   - Check specifically:
      - Face symmetry and proportions
      - Eye/nose/mouth alignment
      - Cheek and jaw contours
      - Head shape naturalness
   - Deduct points for:
      - 2 points: Each severe facial feature distortion
      - 1 point: Each unnatural proportion
      - 1 point: Each perspective warping artifact
      - 0.5 points: Each subtle shape inconsistency
      
3. Overall Quality (1-10):
   - Start from the average of other scores, then deduct:
      - 1 point: For each category scoring below 6
      - 2 points: For each category scoring below 4
      - 1 point: For combined line and shape issues
   - Deduct additional points for:
      - Overall unnatural appearance
      - Professional portrait usability concerns
      - Any noticeable artifacts or inconsistencies

Return the evaluation in JSON format with the following structure:
{
   "line_straightness": score,
   "shape_congruence": score, 
   "overall_quality": score,
   "detailed_analysis": "List ONLY problems found and justify each deduction"
}

REMINDER:
- Focus exclusively on problems and issues
- Do not mention any positive aspects
- Be extremely critical in evaluation
- Document every deduction
- Overall quality must reflect worst aspects
\end{lstlisting}
\end{multicols}
\end{promptbox}

\\

\midrule

\textbf{Output Results} \\
- \textbf{Line Straightness:} \emph{7.5} \\
- \textbf{Shape Congruence:} \emph{8.5} \\
- \textbf{Overall Quality:} \emph{7.0} \\
- \textbf{Detailed Analysis:} \emph{The image shows a slightly curved line in the background, which is a minor deviation from perfect straightness. The overall quality is slightly lower than average due to the slightly unnatural appearance of the subject's face. The subject's eyes are slightly off-center and the nose is slightly askew, which affects the overall shape congruence. The image is usable for professional portrait purposes, but the overall quality could be improved with some adjustments.} \\

\bottomrule
\end{tabular}
\caption{An example of LineACC-LLM and ShapeACC-LLM evaluated by Qwen-VL.}
\label{tab:LLM}
\end{table*}

\subsection{More Qualitative Results}
\label{sec:app_more}
In this section, we demonstrate more qualitative results compared to other methods. \cref{fig:compare1_3} shows the comparison for portrait correction, rectified wide-angle image rectangling, and super-resolution.

From the qualitative visual results \cref{fig:compare1_3} and quantitative comparison results \blux{Table 1}, our proposed method achieve competitive results on four tasks in one common structure.
In portrait correction, it is likely attributed to the estimation errors that we merely attained sub-optimal performance on the original LineACC and ShapeACC metrics. However, we achieved the best performance in terms of LineACC-LLM and ShapeACC-LLM.
We obtain obvious better correction results as seen in \blux{Figure 3.1} and \cref{fig:compare1_3}(1). 
In rectified wide-angle image rectangling, the proposed method improves the PSNR by 1dB and the SSIM by 0.3. 
Although it does not achieve state-of-the-art results in perceptual metrics, this is primarily due to the presence of blur in the ground truth, rendering it unsuitable for image restoration.
For stitched image rectangling, our method outperforms others on three metrics: PSNR, FID, and LPIPS, showing only a marginal improvement of 0.02 on the SSIM when compared to the best results. Moreover, our visual results exhibit greater detail and fewer differences compared to other methods shown in \cref{fig:compare4}. 
\cref{fig:compare5} demonstrates the visual comparison of rotation correction. The proposed method achieves the best performance as shown in \blux{Table 1}. Moreover, the difference map between our results and the ground truth shows fewer errors compared to other methods.

\subsection{Real-World Scene Results}
\label{sec:cross}
For the data utilized in the paper, our framework can also be applied on real-world test images. 
Considering that the portrait correction dataset \cite{tan2021practical} is real-world and the remaining tasks are based on synthetic data, we focus on testing (1) rectified wide-angle image rectangling, (2) stitched image rectangling, and (3) rotation correction under real-world scenes as depicted in \cref{fig:cross}. 

To evaluate the generalization ability for rectified wide-angle image rectangling, we collect rectified wide-angle image results by recent rectification methods \cite{yang2021progressively, liao2021adeep}, which includes various real-world wide-angle lenses. \cref{fig:cross}.1 shows that our method can also be well adapted to these real-world images. The original wide-angle images have obvious distortion while our results have reduced distortion and keep the rectangular borders.

For stitched image rectangling, we collect typical image stitching dataset \cite{zaragoza2013projective, zara2014} and apply different image stitching methods \footnote{\url{https://www.microsoft.com/en-us/research/product/computational-photography-applications/}} to obtain panorama with seamless high-resolution. After getting the stitched images, we then perform rectangling as seen in \cref{fig:cross}.2. We have acheived the best rectangling visual effects with fewer distortions.

For rotation correction, we directly perform our method on the images from MS-COCO \cite{lin2014microsoft} and the corrected results are shown in \cref{fig:cross}.3. These images have different resolutions and light conditions. However, our method can still perform well.

\subsection{Comparisons and Limitations}
\label{sec:limitation}
Recent work MOWA \cite{liao2024mowa} seems similar to this work. However, our motivations and structures differ in several ways: (1) Different tasks: Our model can perform all the tasks their model addresses, but their model does not consider rotation correction. (2) Unified framework: We provide a mathematical formulation of the general distortion model, whereas they do not offer any definition. (3) Network Architecture: Our network is aimed at our proposed general distortion model and based on up-to-date Mamba blocks.  

The limitations of our work focus on the computations, which are computationally higher than some task-specific models (except diffusion models and some super-resolution networks). It would be attractive for users with limited resources if we could make the UniRect lighter. Furthermore, performance on certain tasks needs to be further improved.

\begin{figure*}[hptb!]
    \centering
    \includegraphics[width=1\linewidth]{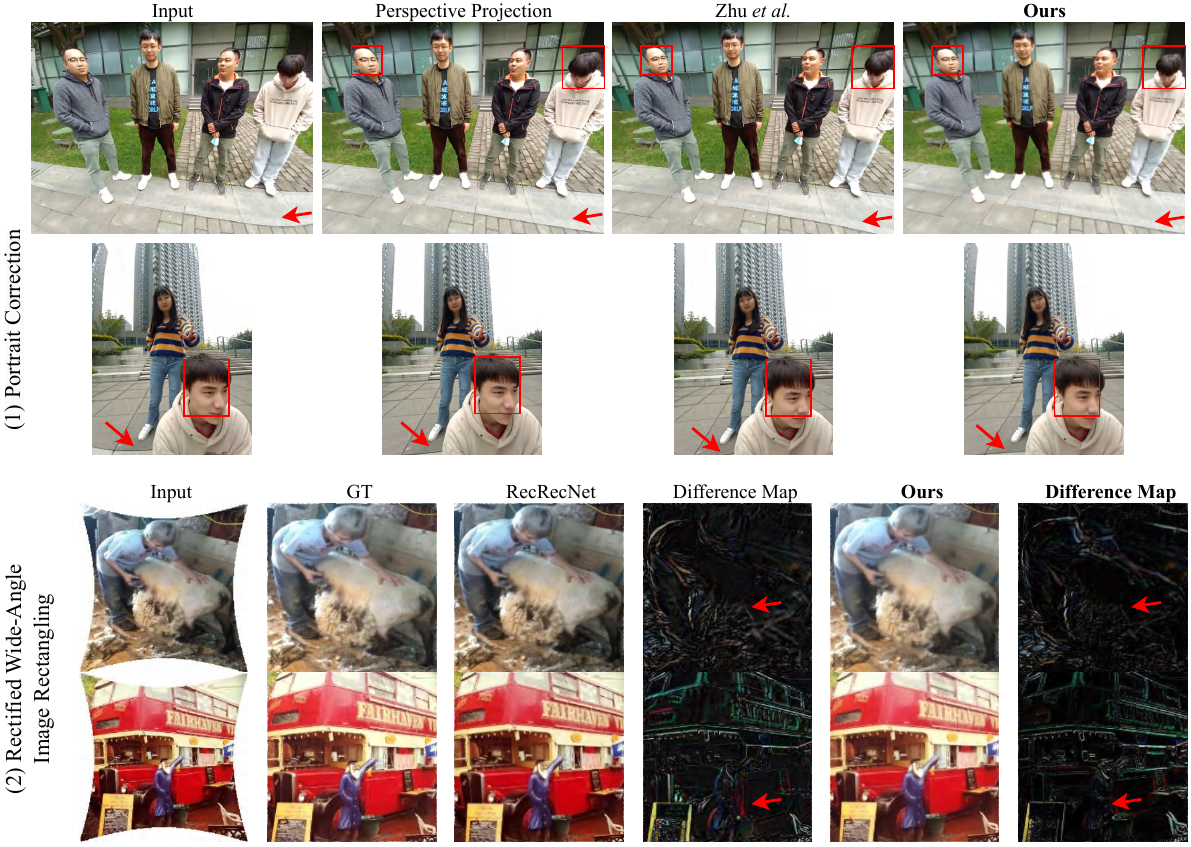}
    \caption{ Qualitative comparison of our UniRect for portrait correction and rectified wide-angle image rectangling. We also give the difference maps between a result image and its groundtruth. Zoom in for best view. }
    \label{fig:compare1_3}
\end{figure*}

\begin{figure*}[htp]
    \centering
    \includegraphics[width=0.95\linewidth]{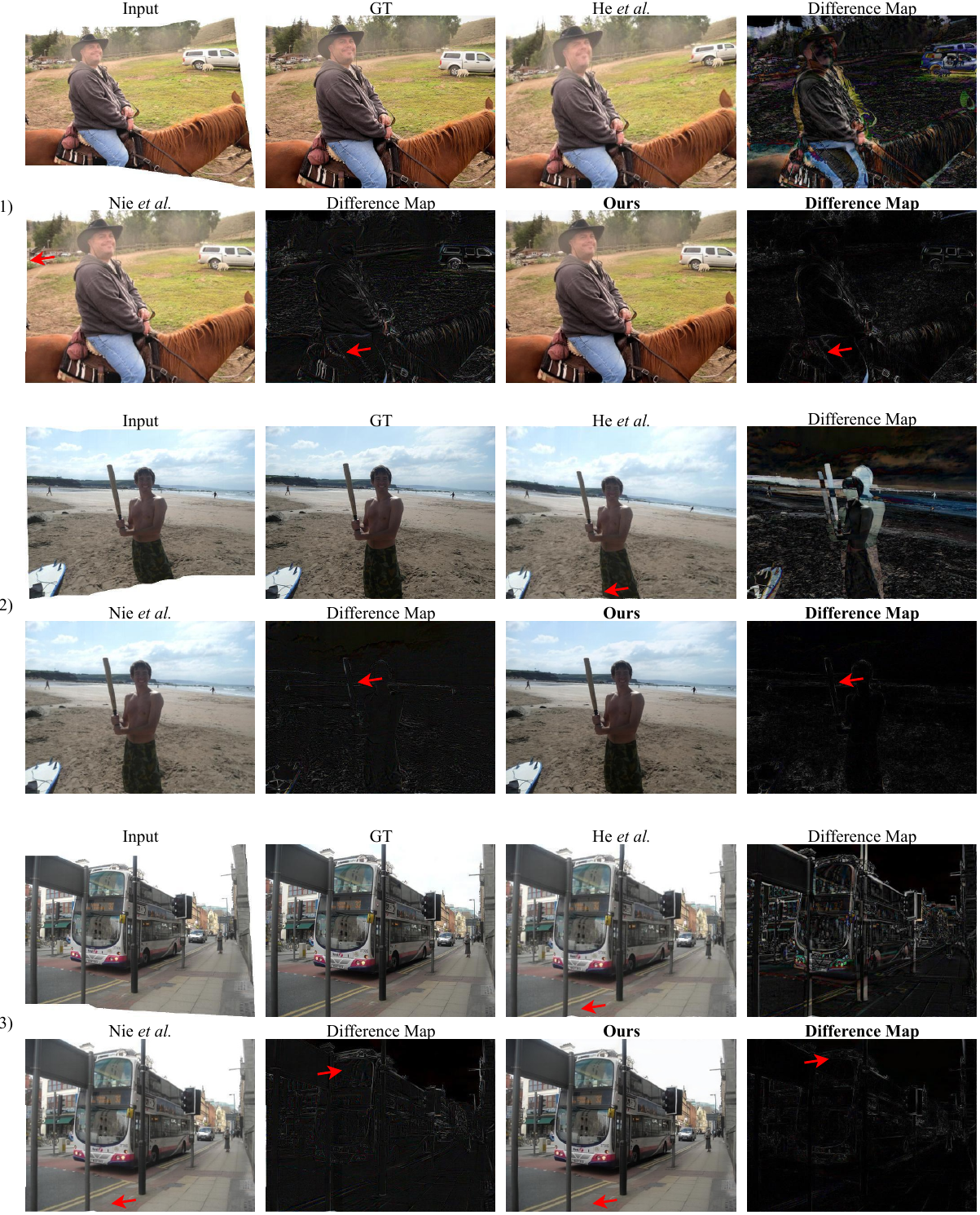}
    \caption{Qualitative comparison of our UniRect for stitched image rectangling. We also give the difference maps between a result image and its groundtruth. Zoom in for best view. }
    \label{fig:compare4}
\end{figure*}

\begin{figure*}[hb]
    \centering
    \includegraphics[width=1\linewidth]{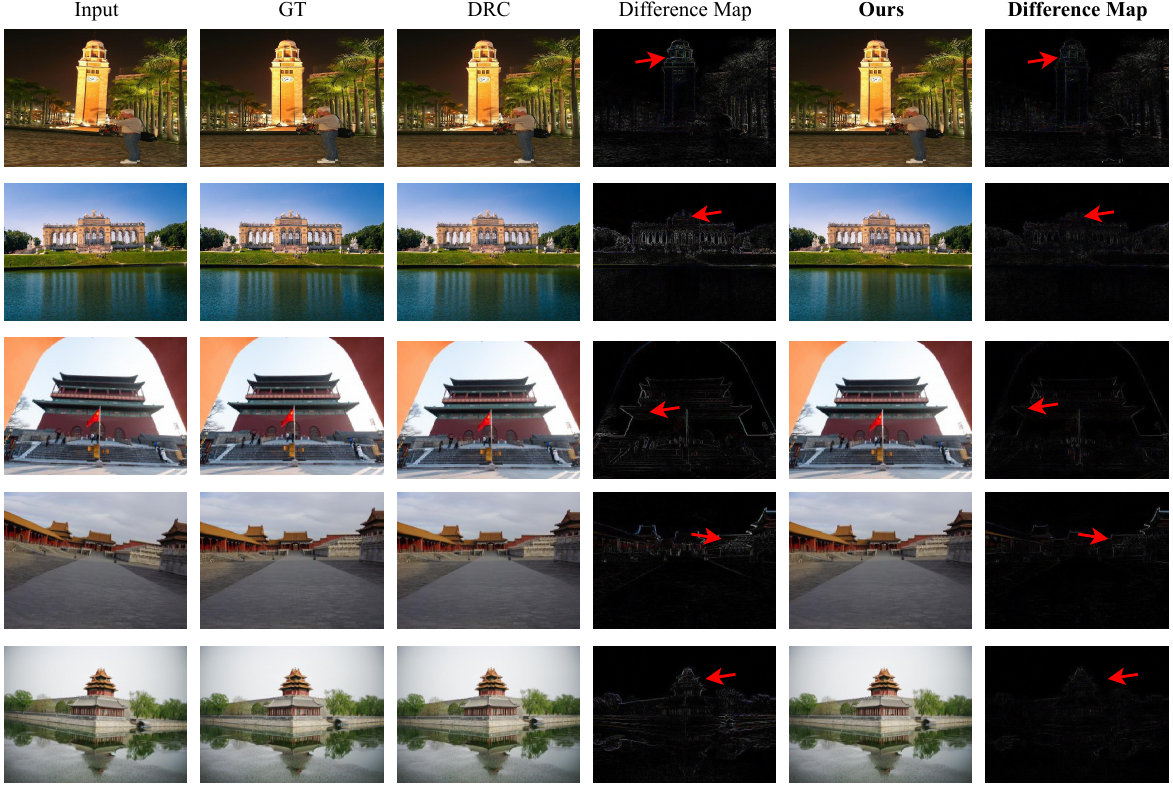}
    \caption{Qualitative comparison of our UniRect for rotation correction. We also give the difference maps between a result image and its groundtruth. Zoom in for best view. }
    \label{fig:compare5}
\end{figure*}

\begin{figure*}[hb]
    \centering
    \includegraphics[width=1\linewidth]{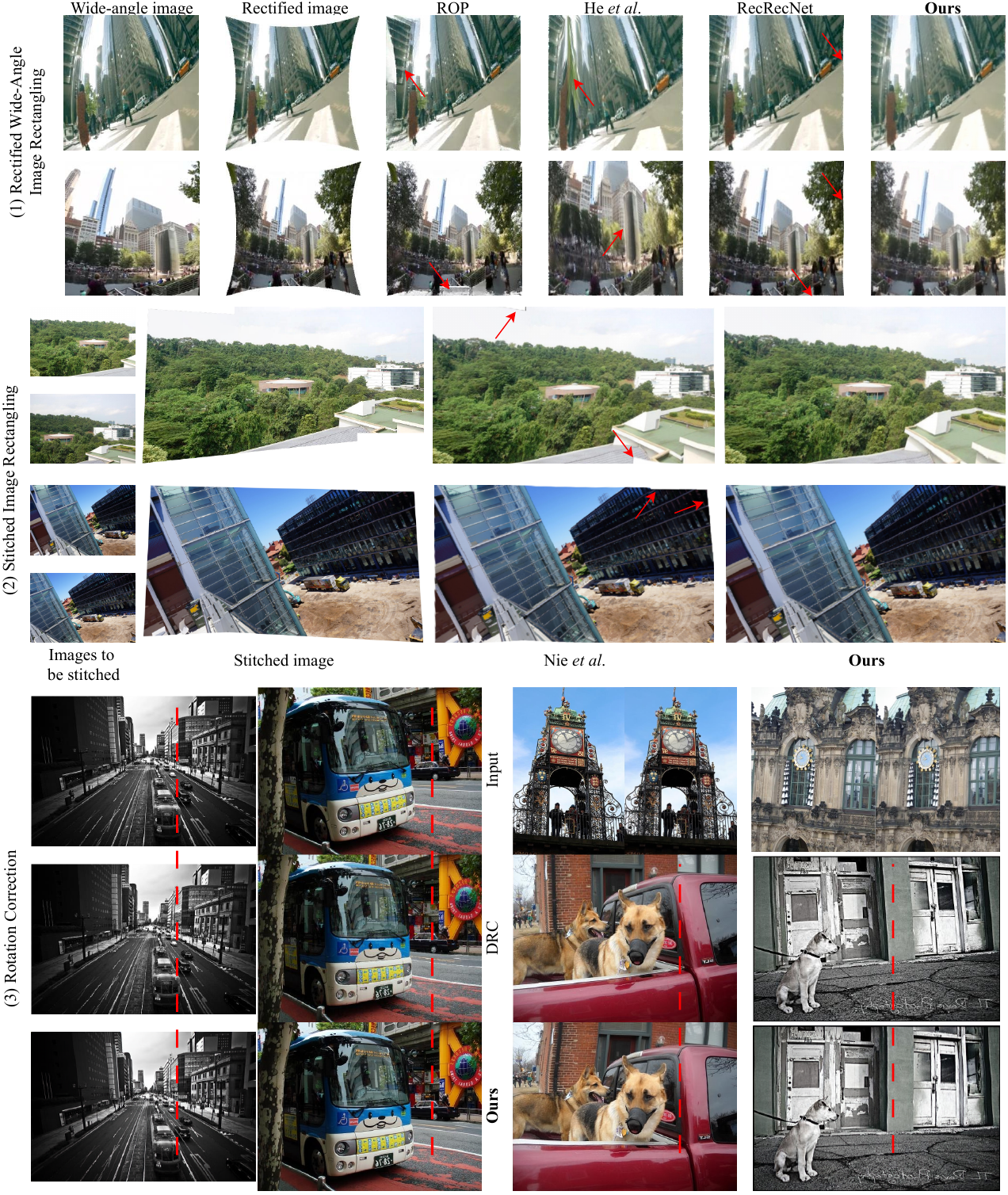}
    \caption{Qualitative comparison of our UniRect for (1) rectified wide-angle image rectangling, (2) stitched image rectangling, and (3) rotation correction under real-world scenes. Zoom in for best view. }
    \label{fig:cross}
\end{figure*}

\end{document}